\pdfoutput=1
\documentclass[11pt,table,xcdraw]{article}
\usepackage[final]{acl}

\usepackage{times}
\usepackage{latexsym}

\usepackage[T1]{fontenc}

\usepackage[utf8]{inputenc}

\usepackage{microtype}

\usepackage{inconsolata}

\usepackage{graphicx}

\usepackage{enumitem}
\usepackage{booktabs}
\usepackage{amsmath}
\usepackage{graphicx}
\usepackage{booktabs}
\usepackage{tabularx}
\usepackage{multirow}
\usepackage{titling}
\usepackage{enumitem}

\usepackage[many]{tcolorbox}

\usepackage{array} 
\usepackage{siunitx}
\usepackage{dcolumn}
\usepackage{xspace}
\usepackage{listings}
\newcommand{\model}[1]{\lstinline[breaklines=true, basicstyle=\ttfamily]|#1|\xspace}

\newcolumntype{L}{D{.}{.}{-3}}

\title{When People are \textit{Floods}: Analyzing Dehumanizing Metaphors in Immigration Discourse with Large Language Models \\
{ \color{red!70!black} \small \textit{Warning: this paper contains examples of upsetting and offensive content.}}}

\author{Julia Mendelsohn \\
  University of Maryland \\
  \texttt{juliame@umd.edu} \\\And
  Ceren Budak \\
  University of Michigan \\
  \texttt{cbudak@umich.edu} \\}

\begin{document}
\maketitle

\begin{abstract}
Metaphor, discussing one concept in terms of another, is abundant in politics and can shape how people understand important issues. We develop a computational approach to measure metaphorical language, focusing on immigration discourse on social media. Grounded in qualitative social science research, we identify seven source domain concepts evoked in immigration discourse (e.g. \textsc{water} or \textsc{vermin}). We propose and evaluate a novel technique that leverages both word-level and document-level signals to measure metaphor with respect to these source domains. We then study the relationship between metaphor, political ideology, and user engagement in 400K US tweets about immigration. While conservatives tend to use dehumanizing metaphors more than liberals, this effect varies widely across source domains. Moreover, creature-related metaphor is associated with more retweets, especially for liberal authors. Our work highlights the potential for computational methods to complement qualitative approaches in understanding subtle and implicit language in political discourse.\footnote{Code and data are available at \url{https://github.com/juliamendelsohn/when_people_are_floods}}

\end{abstract}
\section{Introduction}

Metaphor, communication of one concept in terms of another, is abundant in political discourse. Rooted in culture and cognition, metaphor structures how we interpret the world \citep{lakoff1980metaphors}, and is deployed consciously and subconsciously to shape our understanding of political issues in terms of accessible everyday concepts \citep{burgers_figurative_2016}. By creating conceptual mappings that emphasize some aspects of issues and hide others \citep{lakoff1980metaphors}, metaphor can affect public attitudes and policy preference \citep{boeynaems_effects_2017}. 

Grounded in linguistics and communication literature, we develop a new computational approach for measuring and analyzing metaphor at scale. We use this methodology to study dehumanizing metaphor in immigration discourse on social media, and analyze the relationship between metaphor use, political ideology, and user engagement.

\begin{figure}
    \centering
    \includegraphics[width=.65\columnwidth]{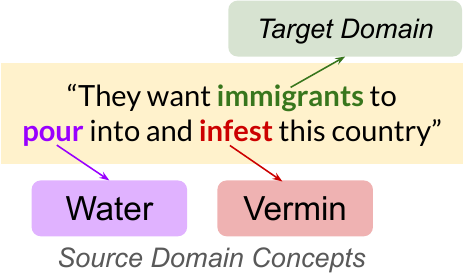}
    \caption{Dehumanizing sentence likening immigrants to the \textit{source domain concepts} of \textsc{water} and \textsc{vermin} via the words ``pour'' and ``infest''.}
    \label{fig:intro}
\end{figure}

From prior discourse analysis literature, we first identify seven \textit{source domains}: concepts evoked in discussions of immigration such as \textsc{water} or \textsc{vermin} (Figure \ref{fig:intro}). We use large language models (LLMs) to detect metaphorical words along with document embeddings to detect metaphorical associations in context. Our method requires no manual annotation, but rather just (1) brief concept descriptions, and (2) a handful of example sentences that evoke each metaphorical concept. We evaluate our approach by creating a new crowdsourced dataset of 1.6K tweets labeled for metaphor, and compare several LLMs and prompting strategies. While we focus on U.S. immigration discourse on social media, our approach can be applied to other political, cultural, and discursive contexts.

We then analyze metaphor usage in 400K U.S. tweets about immigration, with a specific focus on the relationship between metaphor, political ideology, and user engagement. We find that conservative ideology is associated with greater use of dehumanizing metaphor, but this effect varies across concepts. Among conservatives, more extreme ideology is associated with higher metaphor use. Surprisingly, while moderate liberals are more likely to use object-related metaphor, extreme liberal ideology is associated with higher use of creature-related metaphor. Moreover, creature-related metaphors are associated with more retweets, and this effect is primarily driven by liberals. We additionally conduct a qualitative analysis to identify diverse contexts in which liberals use such dehumanizing metaphor. Our study reveals nuanced insights only made possible by our novel approach, and highlights the importance and complexity of studying metaphor as a rhetorical strategy in politics.

\section{Background}

According to Conceptual Metaphor Theory, ``the essence of metaphor is understanding and experiencing one thing in terms of another'' \cite[p.~5]{lakoff1980metaphors}. Metaphor helps people understand complex issues in terms of more concrete everyday experiences \citep{burgers_figurative_2016}. Because metaphors highlight some aspects of issues while downplaying others, they are a type of framing device that can affect how people interpret political issues, with implications for policy recommendation and political action \citep{lakoff1980metaphors,Entman1993,burgers_figurative_2016}.

\paragraph{Immigration Metaphor \& Ideology}

There is a large body of critical discourse scholarship focused on the metaphorical framing of immigration, particularly in news media. Metaphors have long been used in communication about immigration as they ``facilitate listeners' grasp of an external, difficult notion of society in terms of a familiar part of life'' \citep{ana_like_1999}. Metaphors of immigrants as \textsc{animals}, \textsc{vermin}, \textsc{objects}, and \textsc{water} have appeared in U.S. immigration debates for centuries \citep{obrien_indigestible_2003,card2022computational}. Metaphorical dehumanization is sometimes overt (e.g. calling immigrants \textit{animals}) \citep{ana_like_1999}, but can be subconscious as some metaphors are conventionalized in immigration discourse (e.g. the \textsc{water} metaphor evoked by \textit{waves of immigration}) \citep{porto_water_2022}. By emphasizing perceived threats from immigrants, dehumanizing metaphors can increase discrimination and harsh immigration policy support \citep{ana_like_1999,utych_how_2018}.

If metaphor is used to promote such policies, we would expect conservatives to use them more than liberals. This would align with conservative Twitter users tending to frame immigrants as threats \citep{mendelsohn2021modeling}, and Republicans' speeches using more dehumanizing metaphor than Democrats' \citep{card2022computational}. However, prior work finds little-to-no differences between left and right-leaning newspapers' use of immigration metaphors \citep{arcimaviciene_migration_2018,benczes_migrants_2022,porto_water_2022}. Some metaphors, especially \textsc{water}, appear across the ideological spectrum, and are even reinforced by pro-immigration authors \citep{el2001metaphors}.

Beyond binary ideology, ideology strength may impact metaphor use. If dehumanizing metaphors are used to communicate threat and advocate for stringent immigration policies, we would expect the highest use for far-right authors. However, both ideological extremes use more negative emotional language than moderates \citep{alizadeh2019psychology,frimer2019extremists}. If metaphors communicate such emotions \citep{ortony1975metaphors}, their use may be higher for both extremes than for moderates. 

\paragraph{Metaphorical Framing Effects}

How metaphor affects audience's attitudes and behaviors remains an open question. Critical discourse analysis asserts that discourse is not just shaped by society, but also actively constructs social realities; from this lens, metaphor inherently has strong effects on social and political systems \citep{charteris-black_britain_2006,boeynaems_effects_2017}. However, quantitative experiments have shown mixed (and sometimes irreplicable) results \citep{thibodeau_metaphors_2011,steen_when_2014,boeynaems_effects_2017,brugman_metaphorical_2019}. Metaphors' effects vary across factors such as topic, conceptual domains, message source, political orientation, and personality \citep{bosman_persuasive_1987,mio_metaphor_1997,robins_metaphor_2000,kalmoe_fueling_2014,kalmoe_mobilizing_2019,panzeri_does_2021}.

There is experimental evidence of immigration metaphor effects: exposure to \textsc{animal} metaphors increases support for immigration restriction, with the effect mediated by emotions of anger and disgust 
\citep{utych_how_2018}, and exposure to the \textsc{water} metaphor increases border wall support \citep{jimenez2021walls}. Framing the U.S. as a \textsc{body} amplifies effects of contamination threat exposure on negative attitudes towards immigrants \citep{landau_dirt_2014}.
Immigration metaphors' effects are moderated by contextual variables such as intergroup prejudices and ideology \citep{marshall2018scurry,mccubbins_effects_2023}. For example, \textsc{animal} and \textsc{vermin} metaphors increase support for for-profit immigration detention centers, but only among participants with anti-Latino prejudice \citep{mccubbins_effects_2023}. Political ideology may also moderate metaphorical framing effects. 
Due to policy positions and greater sensitivity to threat and disgust \citep{jost2003political,inbar2009conservatives}, effects may be stronger among conservatives. However, prior work finds that liberals are more susceptible to metaphors' effects \citep{thibodeau_metaphors_2011,hart_riots_2018}. Moreover, \citet{lahav2012ideological} find that threat framing has stronger effects on immigration policy attitudes for liberals compared to conservatives. \citet{hart_riots_2018} makes an argument for \textit{entrenchment}: conservatives have more fixed attitudes and opinions, ``while liberals formulate their views on a more context-dependent basis taking into account local information supplied by texts.''

Additionally, recent work has uncovered \textit{resistance to extreme metaphors} among conservatives. Republicans are more opposed to immigrant detention centers when exposed to dehumanizing metaphors \citep{mccubbins_effects_2023}. \citet{hart_animals_2021} find that both \textsc{animal} and \textsc{war} metaphors decrease support for anti-immigration sentiments and policies. \citet{boeynaems_attractive_2023} show that \textsc{criminal} metaphors of refugees push right-wing populist voters' opinions \textit{away} from right-wing immigration stances. Overtly inflammatory metaphors are consciously recognized by audiences, which lead to greater scrutiny of the underlying message \citep{hart_animals_2021}. Due to liberals' sensitivity to metaphor and conservatives' resistance to extreme metaphor, liberals may be more susceptible to the effects of dehumanizing metaphors of immigration. Our work investigates if metaphorical framing effects occur outside of lab experiments by considering user engagement with real-world messages on social media.

\paragraph{Computational Metaphor Analysis}

Metaphor processing encompasses detection, interpretation, generation, and application tasks \citep{shutova_models_2010,rai_survey_2020,ge_survey_2023,kohli_cracking_2023}. Most NLP work focuses on detection as a word-level binary classification task \citep{birke2006clustering,steen2010method,mohler2016introducing}, using linguistic features \citep{neuman_metaphor_2013,tsvetkov_metaphor_2014,jang_metaphor_2016}, neural networks \citep{gao_neural_2018,mao_end--end_2019,dankers_being_2020,le_multi-task_2020}, and BERT models  \citep{liu_metaphor_2020,choi_melbert_2021,lin_cate_2021,aghazadeh_metaphors_2022,babieno_miss_2022,li_framebert_2023,li_finding_2024}. Recent work explores detection and generation with LLMs \citep{dankin_can_2022,liu_testing_2022,joseph_newsmet_2023,lai_multilingual_2023,prystawski_psychologically-informed_2023,ichien_large_2024}. Metaphor detection can also support related tasks, such as propaganda, framing, and hate speech detection \citep{huguet_cabot_pragmatics_2020,lemmens_improving_2021,baleato_rodriguez_paper_2023}.

Few NLP studies examine political metaphor. A study of U.S. politicians' Facebook posts finds that metaphor is associated with higher audience engagement \citep{prabhakaran_how_2021}. NLP work on dehumanization highlights metaphors such as \textsc{vermin}, using embedding-based techniques to quantify such associations \citep{mendelsohn2020framework,engelmann_dataset_2024,zhang_beyond_2024}. \citet{sengupta-etal-2024-analyzing} extend a dataset of news editorials and persuasiveness judgments with metaphor annotations, and find that liberals (but not conservatives) judge more metaphorical editorials to be more persuasive, with effects varying across source domains. Focusing on migration, \citet{zwitter_vitez_extracting_2022} create a dataset of metaphors in Slovene news, with \textsc{water} being the most prevalent source domain. \citet{card2022computational} use BERT token probabilities to quantify associations between immigrants and non-human entities, finding that in recent decades, Republicans have been more likely than Democrats to use such metaphors in political speeches. 

Guided by the aforementioned scholarship, we put forth four research questions and hypotheses:
\footnote{For brevity, we use the term ``metaphor'' to refer to metaphorical dehumanization of immigrants.}
 \begin{itemize}[noitemsep,topsep=0pt]
    \item \textbf{H1}: Conservative ideology is associated with greater metaphor use.
    \item \textbf{RQ1}: Is extreme ideology more associated with metaphor than moderate ideology? 
    \item \textbf{H2}: Higher metaphor use is associated with more user engagement.
    \item \textbf{RQ2}: How does ideology moderate the relationship between metaphor and engagement?
\end{itemize}

\begin{table}[t!]
\centering
\resizebox{.9\columnwidth}{!}{%
\begin{tabular}{@{}ll@{}}
\toprule
Source Domain &
  Example Expressions  \\ \midrule
\textsc{animal} &
  \begin{tabular}[c]{@{}l@{}}sheltering and feeding refugees\\ flocks, swarms, or stampedes\end{tabular}  \\ \midrule
\textsc{vermin} &
  \begin{tabular}[c]{@{}l@{}}infest or plague the country\\ crawling or scurrying in\end{tabular} \\ \midrule
\textsc{parasite} &
  \begin{tabular}[c]{@{}l@{}}leeches, scroungers, freeloaders\\ bleed the country dry\end{tabular}  \\ \midrule
\begin{tabular}[c]{@{}l@{}}\textsc{physical}\\\textsc{pressure}\end{tabular} &
  \begin{tabular}[c]{@{}l@{}}country bursting with immigrants\\ crumbling under the burden \end{tabular}  \\ \midrule
\textsc{water} &
  \begin{tabular}[c]{@{}l@{}}floods, tides, or waves \\ pouring into the country\end{tabular}  \\ \midrule
\textsc{commodity} &
  \begin{tabular}[c]{@{}l@{}}migrants as a cheap source of labor\\ being processed at the border\end{tabular}  \\ \midrule
\textsc{war} &
  \begin{tabular}[c]{@{}l@{}}immigrants as an invading army\\ hordes of immigrants marching in\end{tabular}  \\ \bottomrule
\end{tabular}%
}
\caption{Selected source domains (metaphorical concepts) for analysis. Appendix Table \ref{tab:concepts} has an expanded version with literature references for each concept.}
\label{tab:concepts_short}
\end{table}

\section{Source domains}
We select seven source domains that have been well-documented in prior discourse analysis literature about immigration: \textsc{animal, vermin, parasite, physical pressure, water, commodity,} and \textsc{war} (Table \ref{tab:concepts_short}). While they all liken immigrants to non-human entities, each source domain creates a distinct logic about the perceived threat and plausible remedies. For example, \textsc{water} and \textsc{physical pressure} suggest that immigrants are a threatening force on the host country, metaphorically represented as a container \citep{charteris-black_britain_2006}. Potential solutions would reinforce the container, e.g., through border security. Organism-related metaphors such as \textsc{vermin} and \textsc{parasite} create conceptual mappings through which the \textit{existence} of immigrants is a threat, thus making extermination and eradication plausible responses \citep{steuter_vermin_2010,musolff_metaphorical_2014,musolff_dehumanizing_2015}. 

Some source domains may be seen as more extreme and blatantly dehumanizing than others (e.g. \textsc{vermin} vs. \textsc{commodity}). There is also heterogeneity within each source domain; for example, \textit{historical waves of immigration} and \textit{immigrants are flooding in} both evoke the \textsc{water} metaphor, but the former may be viewed as more conventional with a more neutral valence. Note that there are relevant source domains for immigration discourse beyond these seven, such as \textsc{plants} \citep{ana_like_1999,goncalves_promoting_2024}, which could be further explored in future work.

\section{Measuring Metaphors}

\begin{figure}[t!]
    \centering
    \includegraphics[width=.9\columnwidth]{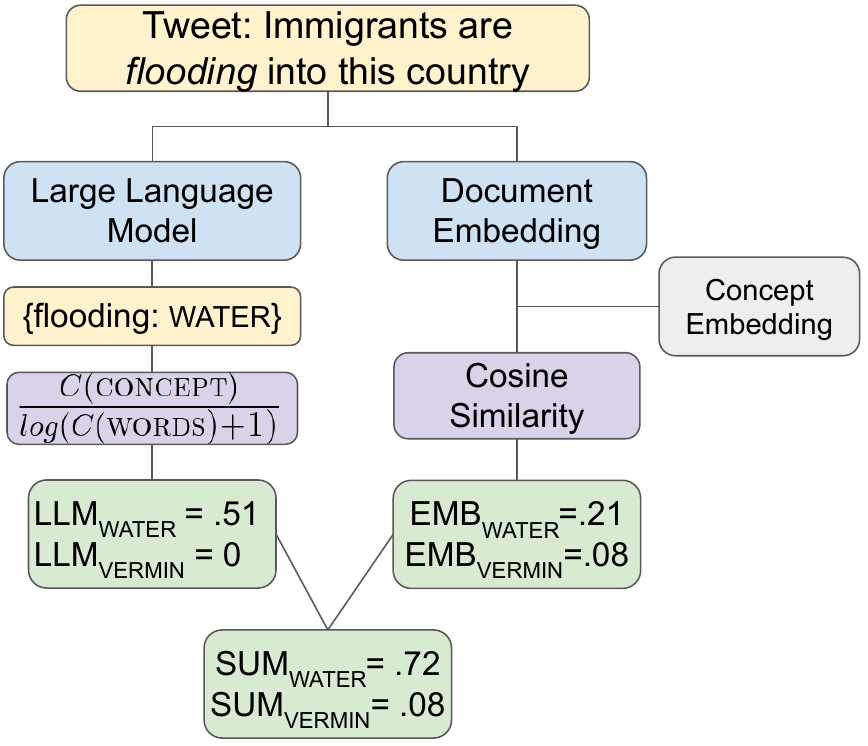}
    \caption{Our method involves calculating the rate of metaphorical words as detected by an LLM (left side), metaphorical associations between documents and source domain concepts with cosine similarity (right side), and adding these measurements together (bottom).}
    \label{fig:metaphor-processing}
\end{figure}

We propose a new approach that accounts for both word- and discourse-level metaphor. Even if a message does not borrow specific words from another source domain, its broader logic could still implicitly be evoked \citep{brugman_metaphorical_2019}. While automatic metaphor processing has primarily focused on lexical information, several computational researchers have argued that broader situational and discourse-level information is necessary to study metaphor production and comprehension \citep{mu2019learning,dankers_being_2020}. We calculate separate word-level and discourse-level measurements using LLMs and document embeddings, respectively, which are then combined to obtain a single \textit{metaphor score} for each source domain (Figure \ref{fig:metaphor-processing}).

\paragraph{Word-level Metaphor Processing}

We first prompt an LLM to identify and map metaphorical expressions to a source domain (or ``none'' if no source domain applies). For example, in Figure \ref{fig:metaphor-processing}, the LLM outputs: \{flooding:\textsc{water}\}. We test two zero-shot prompts. The \texttt{Simple} prompt gives basic instructions and concept names. The \texttt{Descriptive} prompt also provides a definition of metaphor and brief concept descriptions. We evaluate three LLMs: \model{Llama3.1-70B}, \model{GPT-4-Turbo-2024-04-09}, and \model{GPT-4o-2024-08-06}.\footnote{All with temperature = 0. See Appendix for full prompts.}

We calculate word-level scores for each source domain as the number of identified metaphorical expressions ($C(\textsc{concept})$), normalized by log-scaled document length ($C(\textsc{words})$): LLM$_{\textsc{concept}} = \frac{C(\textsc{concept})}{log(C(\textsc{words})+1)}$. Figure \ref{fig:metaphor-processing} has 1 \textsc{water} metaphor in a 6-word tweet, yielding a \textsc{water} score (LLM$_{\textsc{water}}=0.51$). There are 0 \textsc{vermin} metaphors, so LLM$_{\textsc{vermin}}=0$.\footnote{We normalize by length since a short tweet that primarily consists of a metaphor is more ``metaphorical'' than a long tweet that includes a metaphorical word. However, this intuition does not hold linearly. For example, an 8-word tweet containing 2 metaphors (.25) is likely not 5x as  metaphorical as a 40-word tweet containing 2 metaphors (.05).}

\paragraph{Discourse-level Metaphor Processing}

In their study of dehumanizing language, \citet{mendelsohn2020framework} estimate associations between target groups and the \textsc{vermin} metaphor as the cosine similarity between word2vec representations of group labels and an average of vermin-related word embeddings. We adopt this approach with several adjustments. First, we use Transformer-based contextualized embeddings instead of static word embeddings. Second, we calculate associations between source domains and documents, rather than target group labels. Prior approaches require documents to contain mentions of a target group \citep{mendelsohn2020framework,card2022computational}, but many relevant social media posts about sociopolitical issues do not explicitly name the impacted social groups.

We encode each tweet and source domain with SBERT using the \model{all-MiniLM-L6-v2} model \citep{reimers2019sentence}. 
Preliminary investigations reveal that just embedding the source domain name (e.g. the word \textit{water}) overemphasizes literal usages (e.g. migrants crossing the sea). To encourage the model to identify metaphorical associations, we represent each source domain as a set of ``carrier sentences'', which resemble metaphorical usages of the concept but contain little additional semantic information (e.g. \textit{they flood in}; \textit{they hunt them down}). We manually construct 104 carrier sentences based on examples from prior literature (Table \ref{tab:concepts}). Each source domain has 8-22 carrier sentences (Table \ref{tab:sentences}). We calculate the discourse-level score with respect to a given source domain (EMB$_{\textsc{concept}}$) as the cosine similarity between the tweet and the average of the carrier sentences' SBERT representations.

We propose a combined score (SUM), simply the sum of the word- and discourse-level scores. SUM tends to put more weight on the presence of metaphorical words. As metaphorical words are relatively sparse, SUM can still leverage discourse-level signals their absence. Other combination strategies may achieve higher performance, but such explorations are left for future work.

\section{Data}

We use the Immigration Tweets Dataset from \citet{mendelsohn2021modeling}, which has 2.6 million English-language tweets from 2018-2019 that contain a keyword related to immigration (e.g. \textit{immigration}, \textit{illegals}, \textit{undocumented}). The dataset does not contain labels for metaphor, but does include inferred labels for issue-generic, issue-specific, and episodic and thematic frames. The dataset further provides data about user engagement and authors, including their inferred location at the country level and inferred political ideology based on social network-based models \citep{compton2014geotagging,Barbera2015}. We select a random sample of 400K tweets whose authors are based in the United States and have an associated ideology estimate and user engagement metrics. 

The Immigration Tweets dataset already includes estimates of authors' political ideology: \citet{mendelsohn2021modeling} use the Bayesian Spatial Following model from \citet{barbera2015tweeting}. Leveraging assumptions of homophily, this model estimates ideology on a continuous scale based on the Twitter following network and does not consider text. The resulting \textit{ideal points} are continuous estimates ranging from approximately -3 (most liberal) to +3 (most conservative), with values near 0 representing political moderates. We construct two variables to separately examine the effects of liberal vs. conservative ideology (sign of the ideal point) and ideological strength (magnitude of the ideal point). 

We create a new dataset for evaluation. Because the prevalence of metaphor is not known a-priori, we select documents by stratified sampling using a baseline model heuristic (\model{GPT-4-Turbo} with a \model{Simple} prompt) (§\ref{sampling}). We sample 200 documents for each of the seven source domains, and 200 documents for \textit{domain-agnostic} metaphor, i.e., metaphorical language independent of any particular concept, for a dataset of 1,600 documents.

Our approach considers metaphor on a continuous scale. However, eliciting continuous judgments from humans for a nuanced task such as metaphor identification is difficult. Instead, we collect binary judgments from many annotators and consider ``ground-truth'' labels to be the fraction of annotators who judge the document as metaphorical. We develop a codebook for identifying metaphor associated with each source domain (§\ref{annotation}). The codebook was pilot-tested by two authors, who independently labeled 80 tweets (10 per concept and 10 for domain-agnostic metaphor) and had inter-annotator agreement of 0.67 (Krippendorff's $\alpha$).

\begin{table*}[ht!]
\centering
\resizebox{\textwidth}{!}{%
\begin{tabular}{@{}llllllllll@{}}
\toprule
                                  & \multicolumn{9}{c}{Metaphoricity Classification Threshold (\% of annotators)}                                                                         \\ 
                                & 10\%          & 20\%          & 30\%           & 40\%          & 50\%           & 60\%           & 70\%           & 80\%           & 90\%  \\ \midrule
SBERT                           & 0.608         & 0.618         & 0.625          & 0.625         & 0.638          & 0.65           & 0.678          & 0.675          & 0.647          \\
Llama3.1 + Simple               & 0.653         & 0.657         & 0.661          & 0.665         & 0.673          & 0.68           & 0.717          & 0.728          & 0.748          \\
Llama3.1 + Simple + SBERT       & 0.684         & 0.692         & 0.702          & 0.702         & 0.713          & 0.719          & 0.76           & 0.77           & 0.781          \\
Llama3.1 + Descriptive          & 0.508         & 0.509         & 0.512          & 0.511         & 0.513          & 0.51           & 0.504          & 0.499          & 0.497          \\
Llama3.1 + Descriptive + SBERT  & 0.615         & 0.626         & 0.635          & 0.634         & 0.651          & 0.656          & 0.676          & 0.668          & 0.639          \\
GPT-4o + Simple                 & 0.661         & 0.682         & 0.681          & 0.691         & 0.703          & 0.706          & 0.726          & 0.753          & 0.782          \\
GPT-4o + Simple + SBERT         &0.688          & 0.713         & 0.715          & 0.722         & 0.732          & 0.737          & 0.763          & 0.786          & 0.795          \\
GPT-4o + Descriptive            & 0.626         & 0.661         & 0.684          & 0.699         & 0.726          & 0.751          & 0.794          & 0.833          & 0.849          \\
GPT-4o + Descriptive + SBERT    & 0.677         & 0.709         & 0.731          & 0.742         & 0.771          & 0.796          & \textbf{0.847} & \textbf{0.869} & \textbf{0.866} \\
GPT-4-Turbo + Simple            & 0.688         & 0.66          & 0.643          & 0.64          & 0.65           & 0.642          & 0.673          & 0.679          & 0.724          \\
GPT-4-Turbo + Simple + SBERT    &\textbf{0.714} &0 .695         & 0.682          & 0.679         & 0.689          & 0.682          & 0.718          & 0.726          & 0.752          \\
GPT-4-Turbo + Descriptive       & 0.648         & 0.672         & 0.702          & 0.72          & 0.748          & 0.781          & 0.802          & 0.828          & 0.845          \\
GPT-4-Turbo + Descriptive + SBERT & 0.695 & \textbf{0.717} & \textbf{0.746} & \textbf{0.76} & \textbf{0.789} & \textbf{0.817}      & 0.844          & 0.859          & 0.859          \\ \bottomrule
\end{tabular}%
}
\caption{ROC-AUC scores over all concepts for each model, prompt, and SBERT inclusion combination with ground-truth classification thresholds set at 10\% intervals. \texttt{(GPT-4-Turbo, Descriptive, SBERT)} has the highest performance for the 20-60\% range, and \texttt{(GPT-4o, Descriptive, SBERT)} has the highest performance for the 70-90\% range, but the difference between these two models is not significantly different.}
\label{tab:auc_threshold}
\end{table*}

We recruit participants via Prolific to annotate all 1,600 documents. To simplify the task, participants are assigned to one source domain (or the domain-agnostic condition), provided with the relevant codebook portion, and asked to label 20 tweets with respect to their given source domain. They are encouraged to make a binary judgment, but can select a third  ``don't know'' option if needed.\footnote{Participants are based in the United States and have completed at least 200 tasks with a $\geq$99\% approval rate. They are paid \$1.60 per task (\$16/hour). We do not reject any responses through Prolific, but filter out ``don't know'' labels and labels from annotators who (1) completed the full task in under three minutes, or (2) gave the same response for all documents.}

On average, we obtain eight annotations per tweet, and the mean metaphor score is 0.347 (see §\ref{annotation} for details about annotators and annotations). The overall inter-annotator agreement is 0.32 (Krippendorff's $\alpha$) and varies across source domains (Fig. \ref{fig:agreement}). While lower than between experts, this agreement is both expected and advantageous for our approach. As metaphor comprehension is closely tied to culture and individual cognition, judgments may vary widely across the population. Such heterogeneity reinforces that metaphor is not a clear binary, supporting our continuous measurement approach. Future cognitive science work could investigate why some documents and source domains elicit more disagreement than others.

\begin{figure}
    \centering
    \includegraphics[width=\columnwidth]{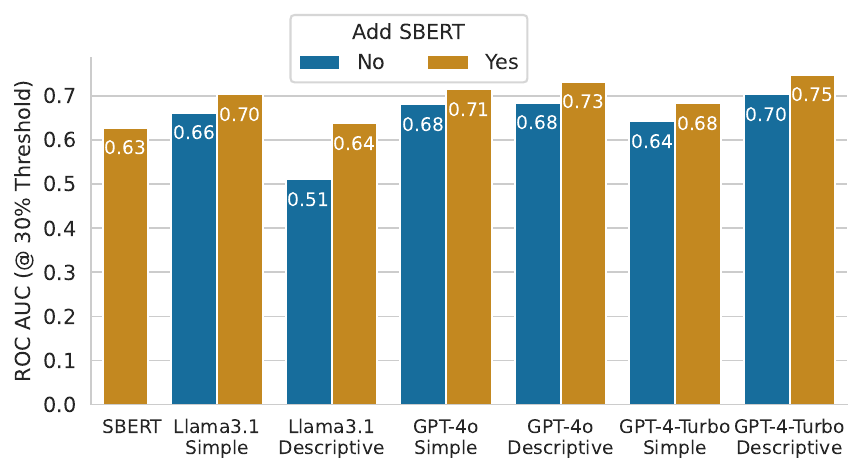}
    \caption{ROC-AUC scores at the 30\% threshold for each LLM and prompt combination, with and without adding discourse-level signal from SBERT. Adding discourse-level metaphorical associations improves performance across all LLMs and prompts.}
    \label{fig:sbert-effect}
\end{figure}

\section{Model Evaluation}

Since both the predicted and ground-truth metaphor scores are continuous, we evaluate models using Spearman correlation and ROC-AUC, applying varied classification thresholds to identify positive instances. In our main analysis, we focus on ROC-AUC values at the 30\% classification threshold (i.e., ``positive instances'' are tweets where at least 30\% of annotators judge it to be metaphorical) because this threshold creates the most balanced dataset. See Appendix §\ref{eval} for complete results.

Table \ref{tab:auc_threshold} and Figure \ref{fig:sbert-effect} show that the best performing models are \model{GPT-4o} and \model{GPT-4-Turbo} with \model{Descriptive} prompts and added \model{SBERT}. These models also have the highest performance across the majority of concepts (Appendix \ref{tab:auc30_concepts}). Notably, including discourse-level signals from adding in SBERT cosine similarity improves performance across all LLMs and prompt combinations.

We measure statistical significance with bootstrap tests (n=100) with a 95\% confidence interval.  \model{GPT-4o} and \model{GPT-4-Turbo} with \model{Descriptive} prompts and added \model{SBERT} outperform all other models, but are not significantly different from each other. We use \model{GPT-4o} for large-scale analysis because inference is 4x cheaper.

\section{Analysis}

We infer metaphor scores for each of the seven source domains for all 400K tweets in our dataset. See Appendix Figures \ref{fig:scores_boxplot}-\ref{fig:scores_by_ideology} for descriptive statistics of metaphor scores. We verify face validity by manually inspecting 50 tweets with highest scores for each source domain, a selection of which are shown in Appendix Table \ref{tab:tweets}, along with the corresponding concepts and authors' ideologies.  

Our analysis uses two sets of regression models. The first quantifies the role of political ideology and ideological strength in metaphor use (§\ref{reg1}), and the second quantifies the association between metaphor, ideology, and user engagement (§\ref{reg2}).

\subsection{Ideology's Role in Metaphor}
\label{reg1}

We quantify the role of ideology in metaphor use with a set of linear regression models. 

\paragraph{Regression Setup} Dependent variables are metaphor scores for each concept. Fixed effects include a binary \textit{ideology} variable (liberal or conservative) and a continuous \textit{ideology strength} score (group-mean centered and z-score normalized), and the interaction between these two variables. These variables capture distinct components of political ideology and enable us to draw conclusions about the far-left, far-right, and moderates.

\begin{figure}
    \centering
    \includegraphics[width=.8\columnwidth]{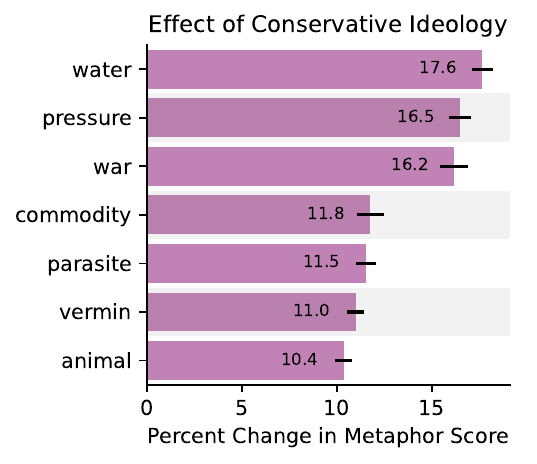}
    \caption{Effect of conservative ideology (relative to liberal ideology) on metaphor score, shown as the average percent change. Percent changes are calculated from marginal effects estimated from regression models.}
    \label{fig:ideology}
\end{figure}

We control for message, author, and time variables (e.g., tweet length, follower count, year and month) as fixed effects. For robustness, we also specify models that control for frames included in the Immigration Tweets Dataset.\footnote{We control for ten issue-generic policy frames\\(e.g., \textit{Security \& Defense}) that were detected by RoBERTa sufficiently well (F1 > 0.6) from \citet{mendelsohn2021modeling}.} See regression tables (\ref{tab:ideology_effect}-\ref{tab:ideology_effect_frames}) for all included variables. We assess significance at the $p=0.05$ level after applying Holm-Bonferroni corrections to account for multiple comparisons. Due to the presence of interaction terms, we enhance interpretability by calculating average marginal effects and visualizing effects as the average percent change in metaphor score.

\begin{figure}
    \centering
    \includegraphics[width=.8\columnwidth]{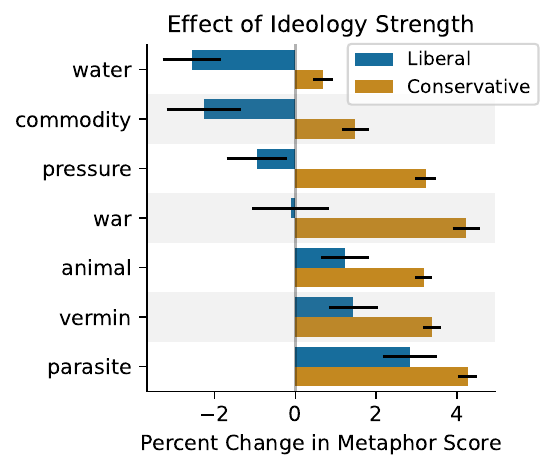}
    \caption{Effect of a one standard deviation increase in ideological strength on metaphor scores. Percent changes are calculated from group-average marginal effects estimated from regression models. Group-average marginal effects reveal that ideological strength has distinct effects on metaphor use among liberals and conservatives, which further vary across concepts.}
    \label{fig:strength}
\end{figure}

\paragraph{Results}

Figures \ref{fig:ideology} and \ref{fig:strength} shows average marginal effects of \textit{conservative ideology} on metaphor scores and group average marginal effects of \textit{ideology strength} for liberals and conservatives, respectively. We visualize marginal effects for ease of interpretability due to the presence of interaction terms. Full regression results are in Appendix Table \ref{tab:ideology_effect}.

\textbf{H1} is supported: conservative ideology is significantly associated with higher scores for all seven concepts. We observe variation across concepts: conservative ideology is most strongly associated with \textsc{war} and \textsc{water}, and least with creature-related metaphors (\textsc{parasite}, \textsc{vermin}, \textsc{animal}).

Addressing \textbf{RQ2}, the relationship between ideology strength and metaphor differs for liberals and conservatives (Fig. \ref{fig:strength}). Among conservatives, extremism is associated with more metaphor across all concepts. For liberals, however, the relationship between strength and metaphor depends on the concept. Extreme liberal ideology is associated with lower use of \textsc{water} and \textsc{commodity} but higher use of creature-related metaphors (\textsc{parasite}, \textsc{vermin}, \textsc{animal}). In sum, both extremes are associated with greater use of creature-related metaphor, but only stronger conservative ideology is associated with greater use of object-related metaphor. These findings hold when controlling for topical frames (Table \ref{tab:ideology_effect_frames} and Figs. \ref{fig:ideology_frames}-\ref{fig:strength_frames}).

\subsection{Metaphor's Role in Engagement}
\label{reg2}

We measure associations between metaphor and user engagement (favorites and retweets) and analyze how effects vary across ideologies.

\paragraph{Regression Setup} We fit linear regression models, where dependent variables are \textit{favorite and retweet counts}, log-transformed as $ln(x+1)$. Independent variables include all concepts' \textit{metaphor scores}, \textit{ideology}, \textit{ideology strength}, and interactions between scores and ideology. We control for the same variables as in §\ref{reg1} (including verified status and follower counts, which strongly predict engagement). We again additionally fit models controlling for issue-generic frames. 

Interpreting the resulting regression coefficients directly is challenging. As in §\ref{reg1}, we calculate average marginal effects, and further back-transform estimates from the $ln(x+1)$ scale to get percent changes in retweet counts. Although the metaphor score is standardized, it is an abstract measure for which one standard deviation difference is not easily interpretable on its own. We thus focus on a ±2 standard deviation range to capture contrasts between tweets with very low and high metaphorical content, providing a more intuitive sense of how metaphorical language relates to user engagement.

\begin{figure}[t!]
    \centering
    \includegraphics[width=.8\columnwidth]{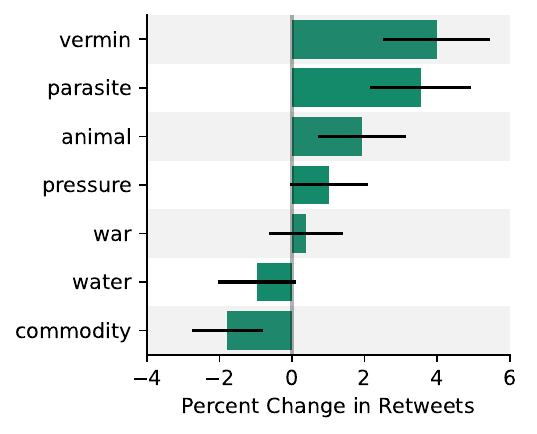}
    \caption{Average marginal effect of metaphor scores on retweets. Effects are shown as percent changes in predicted retweets between non-metaphorical and highly-metaphorical tweets (±2 standard deviations).}
    \label{fig:retweet}
\end{figure}

\begin{figure}[t!]
    \centering
    \includegraphics[width=.8\columnwidth]{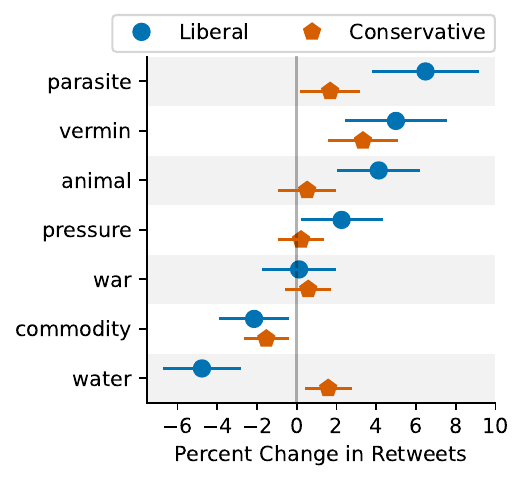}
    \caption{Group-average marginal effects of metaphor on retweets by ideology. Percent changes in predicted retweets between non-metaphorical and highly-metaphorical tweets (±2 standard deviations) are shown.}
    \label{fig:retweet_ideology}
\end{figure}


\paragraph{Results}

Aligned with prior evidence that source domains moderate metaphors' effects \citep{bosman_persuasive_1987}, associations between metaphor scores and user engagement vary by source domain (Fig. \ref{fig:retweet}). Creature metaphors (\textsc{vermin}, \textsc{parasite}, and \textsc{animal}) are associated with more retweets, with the largest effects among liberals (Fig. \ref{fig:retweet_ideology}). Only \textsc{parasite} is significantly associated with more favorites, and metaphors from some domains (e.g. \textsc{commodity}) are actually associated with fewer favorites (Appendix Figs. \ref{fig:favorite}-\ref{fig:favorite_ideology}). For both favorites and retweets, the direction of the effect diverges for only one concept: \textsc{water}, which is associated with higher engagement for conservatives, but lower for liberals. See Appendix Figs. \ref{fig:retweet_frames}-\ref{fig:retweet_ideology_frames} for results when controlling for frames, Tables \ref{tab:engagement}-\ref{tab:engagement_frames} for full regression coefficients, and Tables \ref{tab:mfx_engagement_no_frames}-\ref{tab:mfx_engagement_with_frames} for marginal effects estimates.

\textbf{H2} is partially supported: creature-related metaphors are linked to higher engagement. Addressing \textbf{RQ2}, the relationship between metaphor and engagement is stronger for liberals than conservatives. Crucially, all regressions reveal that relationship between metaphor, ideology, and engagement depends on metaphors' source domains.

\subsection{How are liberals using metaphors?}

Our quantitative analysis paints a complex picture for liberals' use of dehumanizing metaphors of immigrants. While conservative ideology is associated with higher metaphor usage across source domains (§\ref{reg1}), liberals use them to a substantial extent (Appendix Fig. \ref{fig:scores_by_ideology}). Furthermore, among liberals, more extreme ideology is associated with higher creature-related metaphor scores, and the positive relationship between creature-related metaphors and higher retweet counts is driven by liberals. We thus conduct a qualitative analysis of 25 liberal tweets with the highest scores for each concept. Examples discussed here are shown in Table \ref{tab:tweets}. We identify four themes: 

\begin{enumerate}[noitemsep,topsep=0pt,leftmargin=*]
\item Straightforwardly embracing metaphors. Liberals describe migrants as \textit{a source of \$\$\$} (\textsc{commodity}) and \textit{a wave} (\textsc{water}), even if they take pro-immigrant stances.

\item Sympathetic framing, particularly to highlight humanitarian concerns. For example, a liberal overtly cues the \textsc{animal} metaphor while lamenting that ``they hunt them like animals, they cage them like animals''. Other sympathetic instances refer to \textit{feeding} and \textit{sheltering} immigrants; while these verbs can be used to talk about humans, they deny agency to the recipient and are thus more associated with animals \citep{tipler_agencys_2014}.

\item Quoting or paraphrasing outpartisans to highlight their use of inflammatory metaphors. For example, liberal tweets with high \textsc{vermin} metaphor scores quote conservative politicians who refer to immigrants as ``rats'' or an ``infestation'', and such tweets are often critical of blatantly dehumanizing metaphors. 

\item Redirecting dehumanizing metaphors from immigrants to outpartisans. For example, liberal tweets redirect the \textsc{war} metaphor from immigrants to ``right-wing forces'' or invoke the \textsc{parasite} metaphor when referring to Melania Trump (Donald Trump's wife and an immigrant) as a ``tick'' and ``blood sucker''.

\end{enumerate}

\section{Discussion}

More than simply rhetorical decor, metaphors construct and reflect a deeper conceptual structuring of human experiences \citep{lakoff1980metaphors}, and are important devices for political persuasion \citep{mio_metaphor_1997}. While metaphor in politicians' speeches and mass media have been long-studied \citep{charteris-black_britain_2006}, far less is known about how metaphor is used by ordinary people on social media. This theoretical gap is largely driven by a methodological one: measuring metaphorical language at scale is a particularly challenging task.

We develop a computational approach for processing metaphor that uses LLMs and document embeddings to capture both word- and discourse-level signals, which we evaluate on a new dataset of 1600 tweets annotated for metaphor with respect to seven concepts. We apply our approach to analyze dehumanizing metaphor in 400K tweets about immigration, and investigate the relationship between metaphor, political ideology, and user engagement. 

Conservative ideology is associated with greater use of dehumanizing metaphors of immigrants, but varies by source domain. 
This variability and somewhat frequent usage among liberals suggests a high degree of conventionalization in which such metaphors are accepted as ``natural'' \citep{el2001metaphors}. Moreover, compared to moderate liberals, far-left ideology is associated with lower use of objectifying metaphor but higher use of creature metaphor. We conjecture that this pattern may be due to creature metaphors evoking stronger emotions, emphasizing the importance for future metaphor research to consider the role of source domain. Finally, we show that creature-related metaphor is linked to more retweets, with the strongest effects for liberal authors. 
If we assume homophily, i.e., that a tweet's author and audience generally share the same political ideology \citep{barbera2015tweeting}, our results align with prior findings that liberals are more susceptible to the effects of metaphor \citep{hart_riots_2018,sengupta-etal-2024-analyzing}. 

We further qualitatively find that liberals use dehumanizing metaphors to express pro-immigration stances, sympathize, report speech from political opponents, and target outpartisans.
While they may not intend to dehumanize, liberals still tacitly reinforce these metaphors as permissible ways to think and talk about immigrants, with potential ramifications for the treatment of immigrants \citep{el2001metaphors}. Future research could expand our methodology to distinguish between such discursive contexts and examine their social consequences.

Our work offers many avenues for future research. 
Future work could adapt and evaluate our methodology for other issues, languages, and cultures. While we curate source domains from social science literature, such resources may not be available for lesser-studied contexts. Future research could explore developing automated methods for \textit{metaphor discovery}, possibly with the aid of external knowledge graphs and lexical resources \citep{mao_metapro_2022,mao_metapro_2023}. Future analysis-oriented work could use NLP-based measurements of metaphorical language in large-scale experiments to precisely quantify the effects of metaphor on emotions, policy preferences, and social attitudes.

\section{Limitations}

We establish a new framework for computational metaphor analysis with conceptual, methodological, analytical, and resource contributions. In light of this large scope, there are many limitations that future work may consider addressing. 

Our method has many components, each of which could be further optimized. To measure conceptual associations, we only test one document embedding model, one similarity metric, and compare documents with hand-crafted ``carrier sentences''; the specific choice of sentences may also affect performance. Our experiments with LLM-based metaphorical word detection and concept mapping are slightly more comprehensive, as we evaluate three LLMs and two different prompts. However, we do not test few-shot approaches or implement any prompt optimization. We simply add together word-level and concept level scores to get a combined score, and show that this combined score outperforms the individual components. However, future work could evaluate different combination strategies or learn optimal linear combination weights on a held-out set.

Our analysis also has limitations. First is the lack of causality: while we control for various confounds in our regressions, we do not evaluate causal assumptions nor intend to make causal claims. Second is the ambiguity around user engagement as a behavioral outcome. People have diverse motivations for favoriting and retweeting content \citep{meier2014more,boyd2010tweet}, so it is unclear precisely what motivates people to engage in these behaviors, and why we observe stronger associations between metaphor and retweets than favorites. 
It is possible that favoriting activity is dampened by negative emotional content conveyed with dehumanizing metaphors, while retweeting reflects the desire to amplify information that communicates threats \citep{mendelsohn2021modeling}. But, it is not possible to evaluate these mechanisms with the available data. Third, we only have data about favorite and retweet \textit{counts}, not \textit{who} is engaging with the content. This limits interpretations of audience susceptibility to metaphor exposure. We motivate and connect our analysis to prior literature by assuming that authors and their audiences share similar ideologies \citep{barbera2015tweeting}, but this assumption may not always hold. 

Our domain of focus---U.S. immigration discourse on Twitter---is worthy of study in its own right. Nevertheless, the present work is limited in generalizability. We urge future work to extend our methods, evaluation, and analysis to other political issues, platforms, countries, and languages.

\section{Ethical Implications}

We hope this work has positive impact in drawing attention to metaphorical dehumanization, an often unnoticed form of xenophobic discrimination. Our analysis reveals that the dehumanization of immigrants is not limited to the political right. Rather, we all have a responsibility to be aware of dehumanizing metaphors and their implicit societal implications, especially in our own language.

Our primary ethical concerns relate to our reporting of dehumanizing metaphors. Even though we clearly do not endorse these dehumanizing metaphors, merely exposing them to annotators and readers risks reinforcing harmful conceptual associations. Even the act of reporting others' use of slurs can still harm members of targeted communities \citep{croom2011slurs}. It remains an open question if reporting dehumanizing metaphor (even to vehemently disagree with their premise) has similar effects as straightforward usage.  

We recruited hundreds of annotators to help us create our evaluation dataset. The study was deemed exempt by the University of Michigan Institutional Review Board (\#HUM00253235) and annotators were fairly compensated at an average rate of \$16/hour. Nevertheless, creating this dataset involved exposing annotators to offensive and hateful social media posts. We attempt to mitigate these harms by flagging the task as sensitive on Prolific, warning participants of its potentially harmful nature, and limiting each task to just 20 tweets, of which only a few are typically overtly hateful.
\section{Acknowledgments}

We would like to thank Dallas Card, Yulia Tsvetkov, Nicholas Valentino, audiences at MPSA and COMPTEXT 2025, and the ARR reviewers and area chair for their helpful feedback and suggestions. J.M. gratefully acknowledges support from the Google PhD Fellowship.

\newpage
\bibliography{custom}
\newpage
\appendix
\setcounter{figure}{0}
\setcounter{table}{0}
\renewcommand{\thefigure}{A\arabic{figure}}
\renewcommand{\thetable}{A\arabic{table}}

\section{Appendix}

This appendix contains additional details about our annotation process (§\ref{annotation}), prompts, model evaluation (§\ref{eval}), and analysis (§\ref{analysis}).

\subsection{Annotation Details}
\label{annotation}

This section includes details about annotator demographics, annotation statistics, the heuristic-based sampling procedure, and the full codebook.

\begin{table}[htbp!]
\resizebox{\columnwidth}{!}{%
\begin{tabular}{@{}l|ccc|c@{}}
 & Liberal & Moderate & Conservative & Total \\ \hline
Female & 223 & 114 & 45 & 382 \\
Male & 102 & 78 & 53 & 233 \\
Prefer not to say & 2 & 0 & 0 & 2 \\ \hline
Total & 327 & 192 & 98 & 617
\end{tabular}%
}
\caption{Annotator Demographics. All annotators are based in the United States. The table shows the number of annotators across ideology and sex categories, as self-reported to Prolific. The mean age is 38.3 (SD=12.7), and 45 annotators are immigrants (7.3\%).}
\label{tab:demographics}
\end{table}

\begin{table}[htbp!]
\resizebox{\columnwidth}{!}{%
\begin{tabular}{@{}lccccc@{}}
\toprule
concept & \begin{tabular}[c]{@{}c@{}}document\\count\end{tabular} & \begin{tabular}[c]{@{}c@{}}annotation\\count\end{tabular} & \begin{tabular}[c]{@{}c@{}}metaphorical\\annotations\end{tabular}  & \begin{tabular}[c]{@{}c@{}}mean\\score\end{tabular}  \\ \midrule
all & 1600 & 12676 & 4421  & 0.347  \\
animal & 200 & 1898 & 567 &  0.311 \\
parasite & 200 & 1637 & 583  & 0.347  \\
vermin & 200 & 1393 & 348 & 0.246  \\
water & 200 & 1535 & 650 &  0.425  \\
war & 200 & 1475 & 520 & 0.350  \\
commodity & 200 & 1646 & 675 &  0.406  \\
pressure & 200 & 1574 & 610  & 0.385  \\
\begin{tabular}[c]{@{}l@{}}domain-\\agnostic\end{tabular} & 200 & 1518 & 468  & 0.307 \\ \bottomrule
\end{tabular} }
\caption{Descriptive statistics for annotated dataset. \textit{Mean score} refers to the average document score per concept, i.e., the proportion of annotators who labeled a document as metaphorical with respect to the concept.}
\label{tab:annotation-stats}
\end{table}

\begin{figure}[htbp!]
    \centering
    \includegraphics[width=\columnwidth]{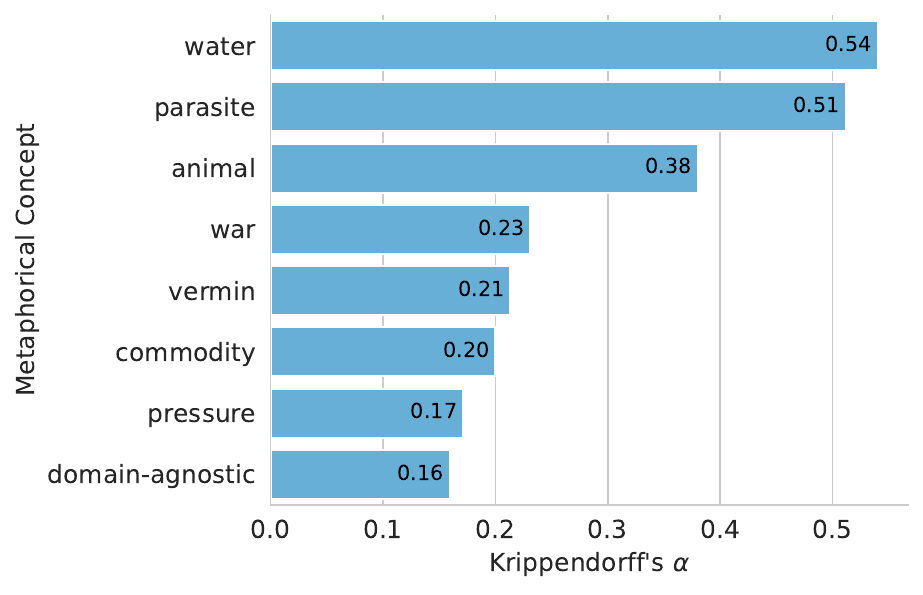}
    \caption{Inter-annotator agreement (Krippendorff's $\alpha$) for each concept and domain-agnostic metaphor.}
    \label{fig:agreement}
\end{figure}

\begin{table*}[htbp!]
\begin{tcolorbox}[colback=white, colframe=white!30!black, boxrule=1pt, arc=4mm, width=\textwidth, boxrule=1pt,title=Codebook,fontupper=\footnotesize]

For each tweet and given concept, label whether or not the tweet evokes metaphors related to the given concept. Focus on the author's language, not their stance towards immigration.\\

To determine if specific words or phrases are metaphorical, consider whether the most basic meaning is related to the listed source domain concept. Basic meanings tend to be more concrete (easier to understand, imagine, or sense) and precise (rather than vague). If you’re not sure about a word’s basic meaning, use the first definition in the dictionary as a proxy. A word should be considered metaphorical if it’s relevant to the listed concept (e.g. animal); it need not exclusively apply to that concept. \\

\textbf{Animal}: Immigrants are sometimes talked about as though they are animals such as beasts, cattle, sheep, and dogs. Label whether or not each tweet makes an association between immigration/immigrants and animals. Label ``YES'' if:
\begin{itemize}[noitemsep, topsep=0pt]
    \item The author uses any words or phrases that are usually used to describe animals. Common examples: \textit{attack}, \textit{flock}, \textit{hunt}, \textit{trap}, \textit{cage}, \textit{breed}
    \item Even if you cannot pinpoint specific words that evoke the concept of animals, if the author's language reminds you of how people talk about animals
\end{itemize}

\textbf{Vermin}: Vermin are small animals that spread diseases and destroy crops, livestock, or property, such as rats, mice, and cockroaches. Vermin are often found in large groups. Label whether or not each tweet makes an association between immigration/immigrants and vermin. Label ``YES'' if:
\begin{itemize}[noitemsep, topsep=0pt]
    \item The author uses any words or phrases that are usually used to describe vermin. Common examples: \textit{infesting}, \textit{swarming}, \textit{dirty}, \textit{diseased}, \textit{overrun}, \textit{plagued}, \textit{virus}
    \item Even if you cannot pinpoint specific words that evoke the concept of vermin, if the author's language reminds you of how people talk about vermin
\end{itemize}

\textbf{Parasite}: Parasites are organisms that feed off a host species at the host’s expense, such as leeches, ticks, fleas, and mosquitoes. Label whether or not each tweet makes an association between immigration/immigrants and parasites. Label “YES” if: 
\begin{itemize}[noitemsep, topsep=0pt]
    \item The author uses any words or phrases that are usually used to describe parasites. Common examples: \textit{leeching}, \textit{freeloading}, \textit{sponging}, \textit{mooching}, \textit{bleed dry}
    \item Even if you cannot pinpoint specific words that evoke the concept of parasites, if the author's language reminds you of how people talk about parasites
\end{itemize}

\textbf{Water}: Immigrants are sometimes talked about using language commonly reserved for water (or liquid motion more broadly). For example, people may talk about immigrants pouring, flooding, or streaming across borders, or refer to waves, tides, and influxes of immigration. Label whether or not each tweet makes an association between immigration/immigrants and water. Label “YES” if: 

\begin{itemize}[noitemsep, topsep=0pt]
    \item The author uses any words or phrases that are usually used to describe water. Common examples: \textit{pouring}, \textit{flooding}, \textit{flowing}, \textit{drain}, \textit{spillover}, \textit{surge}, \textit{wave}
    \item Even if you cannot pinpoint specific words that evoke the concept of water, if the author's language reminds you of how people talk about water
\end{itemize}

\textbf{Commodity}: Commodities are economic resources or objects that are traded, exchanged, bought, and sold. Label whether or not each tweet makes an association between immigration/immigrants. Label ``YES'' if:
\begin{itemize}[noitemsep, topsep=0pt]
    \item The author uses any words or phrases that are usually used to describe commodities. Common examples: \textit{sources of labor}, \textit{undergoing processing}, \textit{imports}, \textit{exports}, \textit{tools}, \textit{being received or taken in}, \textit{distribution}
    \item Even if you cannot pinpoint specific words that evoke the concept of commodities, if the author's language reminds you of how people talk about commodities
\end{itemize}

\textbf{Pressure}: Immigration is sometimes talked about as a physical pressure placed upon a host country, especially as heavy burdens, crushing forces, or bursting containers. Label whether or not each tweet makes an association between immigration/immigrants and physical pressure. Label “YES” if: 
\begin{itemize}[noitemsep, topsep=0pt]
    \item The author uses any words or phrases that are usually used to describe physical pressure. Common examples: host country \textit{crumbling}, \textit{bursting}, being \textit{crushed}, \textit{stretched thin}, or \textit{strained}, immigrants as \textit{burdens}.
    \item Even if you cannot pinpoint specific words that evoke the concept of pressure, does the author’s language remind you of how people talk about physical pressure? 
\end{itemize}

\textbf{Commodity}: Commodities are economic resources or objects that are traded, exchanged, bought, and sold. Label whether or not each tweet makes an association between immigration/immigrants. Label ``YES'' if:
\begin{itemize}[noitemsep, topsep=0pt]
    \item The author uses any words or phrases that are usually used to describe commodities. Common examples: \textit{sources of labor}, \textit{undergoing processing}, \textit{imports}, \textit{exports}, \textit{tools}, \textit{being received or taken in}, \textit{distribution}
    \item Even if you cannot pinpoint specific words that evoke the concept of commodities, if the author's language reminds you of how people talk about commodities
\end{itemize}

\textbf{War}: People sometimes talk about immigration in terms of war, where immigrants are viewed as an invading army that the host country fights against. Label whether or not each tweet makes an association between immigration/immigrants and war. Label “YES” if:
\begin{itemize}[noitemsep, topsep=0pt]
    \item The author uses any words or phrases that are usually used to describe war. Common examples: \textit{invasion}, \textit{soldiers}, \textit{battle}, \textit{shields}, \textit{fighting}
    \item Even if you cannot pinpoint specific words that evoke the concept of war, if the author's language reminds you of how people talk about war
\end{itemize}

\textbf{Domain-Agnostic}: Label whether or not each tweet uses metaphorical (non-literal) language to talk about immigration/immigrants. Metaphorical language involves talking about immigration/immigrants in terms of an otherwise unrelated concept. For example, \textit{waves of immigration} is metaphorical because the word \textit{waves} is associated with water.

\end{tcolorbox}
\end{table*}

\subsubsection{Sampling for Annotation}
\label{sampling}

The prevalence of metaphorical language with respect to each source domain concept is not known a-priori. Instead of randomly sampling tweets for human annotation, we thus adopt a stratified sampling approach using scores from a baseline model (\texttt{GPT-4-Turbo, Simple Prompt}) as a heuristic. 

We get heuristic scores from the baseline model for a set of 20K documents, which we call $D$. For each concept $c$, we sample $n_c$ documents from $D$. Let $h_c$ be the heuristic metaphor score with respect to $c$. $Q_{0,c} \in D$ is then the set of documents with $h_c = 0$. $Q_{i,c} \in D$ where $i = 1,2,...,k-1$ are the $k-1$ quantiles of documents with $h_c > 0$. The annotation sample for each concept is then: $$S_c = Sample(Q_{0,c},\frac{n_c}{k}) \cup  \bigcup_{i=1}^{k-1}Sample(Q_{i,c},\frac{n_c}{k})$$ Using $k=5$ strata, we sample $n_c=200$ tweets for each source domain. We additionally sample 200 documents for \textit{domain-agnostic} metaphor.

Below are examples of tweets from different strata for the \textsc{water} concept:
\begin{itemize}[noitemsep,topsep=0pt]
    \item $Q_{0,\textsc{water}}$: \textit{How about we help US citizens with cancer before spending money on illegals} 
    \item $Q_{2,\textsc{water}}$: \textit{Tough reading. A report on some of the facilities to which the migrant children are shipped after the American government abducts them from their parents. This despicable practice is a permanent stain on the US}
    \item $Q_{4,\textsc{water}}$: \textit{An overwhelming flood of illegal aliens for the middle class to pay for.}

\end{itemize}

\begin{tcolorbox}[colback=blue!10!white, colframe=blue!80!black, boxrule=1pt, arc=4mm, width=\linewidth, boxrule=1pt,title=Simple Prompt]
    For each metaphorical word in the tweet below, select the most relevant concept from the following list:\\
    \verb|[water, commodity, physical pressure,| 
    \verb|war, animal, vermin, parasite]|\\
    Respond with a JSON object where keys are metaphors and values are relevant concepts. \\
    If a metaphor is not related to any concept above, set its value to "none". \\
    If there are no metaphors, output an empty JSON object.
    \\\\
    Tweet: [\verb|TWEET TEXT]|
\end{tcolorbox}

\begin{tcolorbox}[colback=violet!10!white, colframe=violet!80!black, boxrule=1pt, arc=4mm, width=\linewidth, boxrule=1pt,title=Descriptive Prompt]
Analyze the tweet below to identify metaphors used to describe immigrants or immigration. In this context, metaphors are words and phrases that are used non-literally and create associations between immigration and other concepts. For each identified metaphor, select the most relevant concept from the following list: 

Concepts (explanations in parentheses):\\
\textbf{Parasite} (organisms that feed off a host species at the host’s expense, such as leeches, ticks, fleas, and mosquitoes)\\
\textbf{Vermin} (small animals that spread diseases or destroy crops, livestock, or property, such as rats, mice, and cockroaches)\\
\textbf{Animal} (living creatures, such as beasts, cows, dogs, sheep, and birds) \\
\textbf{Water} (or liquid motion more broadly)\\
\textbf{Physical Pressure} (destructive physical force, such as heavy burdens, crushing forces, and bursting containers) \\
\textbf{Commodity} (economic resources or objects that are traded, exchanged, bought, or sold)\\
\textbf{War} (or fights and battles more broadly)\\
Provide your analysis as a JSON object where keys are the metaphors and values are their most relevant concepts. Only include the concept name (e.g. commodity, animal, parasite). Do not include the concept explanation in your response. If a metaphor is not related to any of the listed concepts, set its value to “none”. If no metaphors are found, return an empty JSON object.
    \\\\
    Tweet: [\verb|TWEET TEXT]|
\end{tcolorbox}

\begin{table*}[htbp!]
\centering
\resizebox{\textwidth}{!}{%
\begin{tabular}{@{}lll@{}}
\toprule
Source Domain &
  Example Expressions &
  Sources \\ \midrule
\textsc{animal} &
  \begin{tabular}[c]{@{}l@{}}hunt down and ferret out immigrants\\ sheltering and feeding refugees\\ flocks, swarms, or stampedes of migrants\end{tabular} &
  \begin{tabular}[c]{@{}l@{}}\citet{obrien_indigestible_2003, arcimaviciene_migration_2018}\\ \citet{ana_like_1999, obrien_indigestible_2003, hart_animals_2021}\\ \citet{steuter_vermin_2010, zwitter_vitez_extracting_2022}\end{tabular} \\ \midrule
\textsc{vermin} &
  \begin{tabular}[c]{@{}l@{}}immigrants infest or plague the country\\ immigrants crawling or scurrying in\\ immigrants as cockroach or rat-like\end{tabular} &
  \begin{tabular}[c]{@{}l@{}}\citet{hart_animals_2021,steuter_vermin_2010}\\ \citet{utych_how_2018,musolff_dehumanizing_2015,obrien_indigestible_2003}\end{tabular} \\ \midrule
\textsc{parasite} &
  \begin{tabular}[c]{@{}l@{}}migrants as host of society's ills\\ leeches, scroungers, freeloaders\\ bleed the country dry\end{tabular} &
  \begin{tabular}[c]{@{}l@{}}\citet{ana_like_1999, markowitz_social_2020} \\ \citet{musolff_metaphorical_2014,musolff_dehumanizing_2015}\end{tabular} \\ \midrule
\textsc{physical pressure} &
  \begin{tabular}[c]{@{}l@{}}immigrants are a burden\\ country bursting with immigrants\\ crumbling under the weight of immigrants\end{tabular} &
  \begin{tabular}[c]{@{}l@{}}\citet{abid_flood_2017,ana_like_1999}\\ \citet{charteris-black_britain_2006,zwitter_vitez_extracting_2022}\end{tabular} \\ \midrule
\textsc{water} &
  \begin{tabular}[c]{@{}l@{}}floods, tides, or waves of immigrants\\ immigrants pouring into the country\\ absorbing immigrants\end{tabular} &
  \begin{tabular}[c]{@{}l@{}}\citet{abid_flood_2017, arcimaviciene_migration_2018}\\ \citet{charteris-black_britain_2006, martin_masters_2022}\\ \citet{porto_water_2022, ana_like_1999, taylor_affordances_2022}\end{tabular} \\ \midrule
\textsc{commodity} &
  \begin{tabular}[c]{@{}l@{}}fairly redistributing refugees\\ migrants are engines of the economy\\ immigrants being processed at the border\end{tabular} &
  \begin{tabular}[c]{@{}l@{}}\citet{arcimaviciene_migration_2018,obrien_indigestible_2003}\\ \citet{de_backer_persuasive_2022,goncalves_promoting_2024}\end{tabular} \\ \midrule
\textsc{war} &
  \begin{tabular}[c]{@{}l@{}}immigrants are an invading army\\ hordes of immigrants marching in\\ host countries are under siege\end{tabular} &
  \begin{tabular}[c]{@{}l@{}}\citet{ana_like_1999, obrien_indigestible_2003, hart_animals_2021}  \\ \citet{de_backer_persuasive_2022, utych_how_2018} \\ \citet{zwitter_vitez_extracting_2022}\end{tabular} \\ \bottomrule
\end{tabular}%
}
\caption{Selected source domains (metaphorical concepts) for analysis.}
\label{tab:concepts}
\end{table*}

\begin{table*}[htbp!]
\centering
\resizebox{\textwidth}{!}{%
\begin{tabular}{@{}llll@{}}
\toprule
Animal & Vermin & Commodity & Water \\ \midrule
They attack them. & They are cockroaches. & They are distributed between them. & They absorb them. \\
They bait them. & They crawl in. & They are the engine of it. & There is a deluge of them. \\
They breed them. & They are dirty. & They exchange them. & They drain it. \\
They are brutish. & They are diseases. & They export them. & They engulf it. \\
They butcher them. & They fester. & They import them. & They flood in. \\
They capture them. & They are impure. & They are instruments. & They flow in. \\
They catch them. & They infect it. & They are instrumental to it. & There is an inflow of them. \\
They chase them down. & They infest it. & They are packed in. & There is an influx of them. \\
They ensnare them. & There is an infestation of them. & They are processed. & They inundate it. \\
They ferret them out. & They overrun it. & They are redistributed between them. & There is an outflow of them. \\
They flock in. & They are a plague. & They accept a share of them. & There is an overflow of them. \\
They hunt them down. & They are poisonous. & They take them in. & They pour in. \\
They lay a trap for them. & They are rats. & They are tools. & They spill in. \\
They lure them in. & They sneak in. & They trade them in. & There is a spillover of them. \\
They round them up. & There is a swarm of them. &  & There is a storm of them. \\
They slaughter them. & They are a virus. &  & They stream in. \\
They slither in. &  &  & There is a surge of them. \\
They trap them. &  &  & They swamp it. \\
They wiggle in. &  &  & There is a swell of them. \\
 &  &  & There is a tide of them. \\
 &  &  & They trickle in. \\
 &  &  & There is a wave of them. \\ \midrule
Parasite & Physical Pressure & War &  \\ \midrule
They bleed it dry. & It bears the brunt of them. & They are an army. &  \\
They are bloodthirsty. & It buckles under them. & They battle them. &  \\
They are a cancer. & They are a burden. & They bludgeon them. &  \\
They leech off them. & They cause it to burst. & They capture them. &  \\
They are parasites. & They bust it. & They are caught in the crosshairs &  \\
They scrounge around. & They crumble it. & They fight them. &  \\
They are scroungers. & They fill it up. & They are invaders. &  \\
They are spongers. & They are a load on it. & There is an invasion of them. &  \\
 & They put pressure on it. & There are regiments of them. &  \\
 & They seal it up. & They shield them. &  \\
 & They are a strain on it. & They are soldiers. &  \\
 & They stretch it thin. & They are warriors. &  \\ \bottomrule
\end{tabular}%
}
\caption{\textit{Carrier sentences} used to create each concept's SBERT representation. Each sentence evokes a metaphorical, rather than literal, sense of each concept but remains as generic as possible.}
\label{tab:sentences}
\end{table*}

\begin{table*}[htbp!]
\resizebox{\textwidth}{!}{%
\begin{tabular}{@{}lcccccccc@{}}
\toprule
                                  & \multicolumn{8}{c}{ROC-AUC @ 30\% Classification Threshold}                                                                                                                      \\ 
                                  & animal         & commodity      & parasite       & pressure       & vermin         & war            & water          & domain-agnostic\\ \midrule
SBERT                             & 0.738          & 0.589          & 0.601          & 0.586          & 0.697          & 0.662          & 0.669          & -              \\
Llama3.1 + Simple                 & 0.709          & 0.581          & 0.697          & 0.538          & 0.613          & 0.699          & 0.786          & \textbf{0.692} \\
Llama3.1 + Simple + SBERT         & 0.786          & 0.619          & 0.727          & 0.581          & \textbf{0.746} & \textbf{0.775} & 0.804          & -              \\
Llama3.1 + Descriptive            & 0.504          & 0.534          & 0.517          & 0.504          & 0.500          & 0.500          & 0.505          & 0.656          \\
Llama3.1 + Descriptive + SBERT    & 0.725          & 0.613          & 0.616          & 0.588          & 0.697          & 0.662          & 0.676          & -              \\
GPT-4o + Simple                   & 0.731          & 0.613          & 0.662          & 0.563          & 0.610          & 0.714          & 0.856          & 0.510          \\
GPT-4o + Simple + SBERT           & 0.806          & 0.642          & 0.706          & 0.606          & 0.740          & 0.767          & 0.861          & -              \\
GPT-4o + Descriptive              & 0.682          & 0.655          & 0.744          & 0.595          & 0.547          & 0.661          & 0.868          & 0.661          \\
GPT-4o + Descriptive + SBERT      & 0.812          & 0.673          & 0.795          & 0.647          & 0.705          & 0.723          & \textbf{0.890} & -              \\
GPT-4-Turbo + Simple              & 0.658          & 0.575          & 0.673          & 0.538          & 0.606          & 0.698          & 0.809          & 0.685          \\
GPT-4-Turbo + Simple + SBERT      & 0.736          & 0.598          & 0.702          & 0.581          & 0.671          & 0.760          & 0.830          & -              \\
GPT-4-Turbo + Descriptive         & 0.747          & 0.691          & 0.762          & 0.635          & 0.599          & 0.649          & 0.795          & 0.620          \\
GPT-4-Turbo + Descriptive + SBERT & \textbf{0.844} & \textbf{0.712} & \textbf{0.801} & \textbf{0.688} & 0.727          & 0.712          & 0.819          & -              \\ \bottomrule
\end{tabular}%
}
\caption{Evaluation for each concept and domain-agnostic metaphor classification, calculated as the ROC-AUC score at the 30\% classification threshold.}
\label{tab:auc30_concepts}
\end{table*}
\begin{table*}[htbp!]
\resizebox{\textwidth}{!}{%
\begin{tabular}{@{}lccccccccc@{}}
\toprule
                                  & \multicolumn{9}{c}{Spearman Correlation}                                                                                                                \\ 
\textbf{}                         & overall        & animal         & commodity      & parasite       & pressure       & vermin         & war            & water          & domain-agnostic \\ \midrule
SBERT                             & 0.260          & 0.452          & 0.272          & 0.147          & 0.189          & 0.328          & 0.367          & 0.297          & -               \\
Llama3.1 + Simple                 & 0.379          & 0.505          & 0.146          & 0.354          & 0.043          & 0.286          & 0.471          & 0.681          & 0.412           \\
Llama3.1 + Simple + SBERT         & 0.411          & 0.603          & 0.241          & 0.376          & 0.125          & 0.409          & 0.561          & 0.645          & -               \\
Llama3.1 + Descriptive            & 0.076          & -0.027         & 0.130          & 0.110          & 0.121          & NaN            & NaN            & 0.066          & 0.397           \\
Llama3.1 + Descriptive + SBERT    & 0.273          & 0.423          & 0.297          & 0.163          & 0.195          & 0.328          & 0.367          & 0.308          & -               \\
GPT-4o + Simple                   & 0.424          & 0.413          & 0.225          & 0.361          & 0.214          & 0.347          & 0.482          & 0.744          & 0.092           \\
GPT-4o + Simple + SBERT           & 0.443          & 0.545          & 0.286          & 0.377          & 0.262          & \textbf{0.428} & \textbf{0.565} & 0.71           & -               \\
GPT-4o + Descriptive              & \textbf{0.529}          & 0.479          & 0.433          & 0.594          & 0.414          & 0.251          & 0.456          & \textbf{0.767} & 0.395           \\
GPT-4o + Descriptive + SBERT      & 0.48           & 0.591          & 0.396          & 0.520          & 0.352          & 0.348          & 0.5            & 0.733          & -               \\
GPT-4-Turbo + Simple              & 0.333          & 0.366          & 0.167          & 0.317          & 0.119          & 0.239          & 0.445          & 0.66           & 0.366           \\
GPT-4-Turbo + Simple + SBERT      & 0.397          & 0.499          & 0.242          & 0.348          & 0.201          & 0.350          & 0.540          & 0.662          & -               \\
GPT-4-Turbo + Descriptive         & \textbf{0.529} & 0.537          & 0.448          & \textbf{0.623} & \textbf{0.419} & 0.376          & 0.436          & 0.656          & 0.321           \\
GPT-4-Turbo + Descriptive + SBERT & 0.504          & \textbf{0.619} & \textbf{0.466} & 0.543          & 0.398          & 0.407          & 0.489          & 0.629          & -               \\ \bottomrule
\end{tabular}%
}
\caption{Spearman correlations between models' predicted metaphoricity and annotators' scores (defined as the percentage of annotators who labeled a document as metaphorical with respect to a specified concept). Across all concepts, \texttt{GPT-4-Turbo/GPT-4o, Descriptive} has the highest performance, but is not statistically different from \texttt{GPT-4-Turbo/GPT-4o, Descriptive, SBERT}. Statistical significance was determined at the $p < 0.05$ level using the Fisher r-to-z transformation.}
\label{tab:spearman}
\end{table*}

\begin{figure}[htbp!]
    \centering
    \includegraphics[width=\columnwidth]{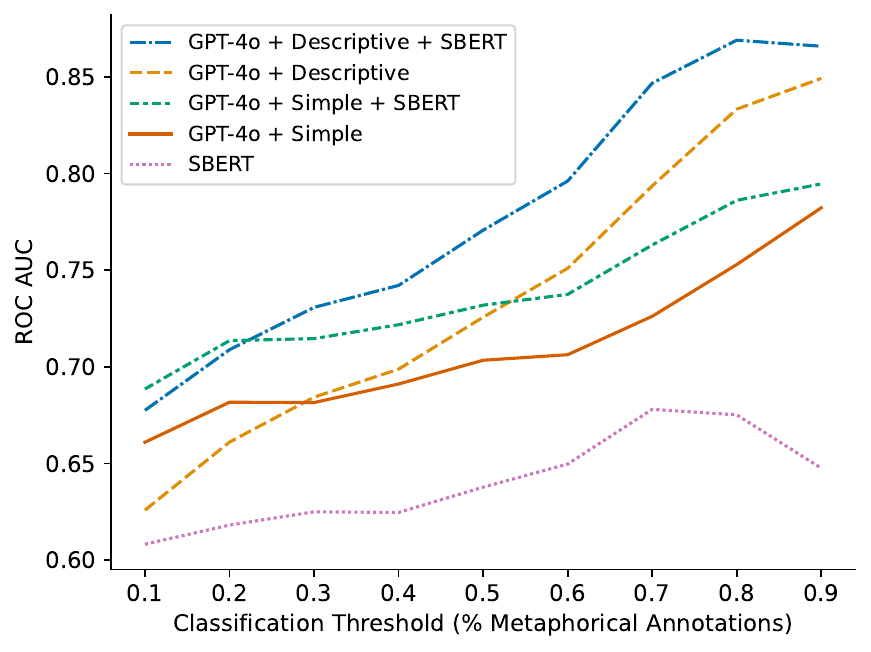}
    \caption{Comparison of GPT-4o-based metaphor scoring models that vary in prompt (\textit{Simple} or \textit{Descriptive}) and whether document-level associations are incorporated with SBERT embeddings. The x-axis represents different classification thresholds (i.e., percent of annotators who label a tweet as metaphorical). Across all thresholds, including SBERT improves performance.}
    \label{fig:auc_by_threshold}
\end{figure}

\subsection{Evaluation}
\label{eval}
Tables \ref{tab:auc30_concepts}-\ref{tab:spearman} shows full model evaluation results across concepts (using ROC-AUC at the 30\% threshold) and Spearman correlations between predicted and annotated scores, respectively. Figure \ref{fig:auc_by_threshold} shows performance at different classification thresholds for \texttt{GPT-4o} approaches. While \texttt{SBERT} on its own has the lowest performance, including SBERT scores in the \texttt{GPT-4o} approaches consistently improves performance across all thresholds.

\begin{table*}[htbp!]
\centering
\resizebox{\textwidth}{!}{%
\begin{tabular}{@{}cccl@{}}
\toprule
Concept &
  Score &
  Ideology &
  Text \\ \midrule
 
  &
  \cellcolor[HTML]{FFE9E8}1.063 &
  \cellcolor[HTML]{FFE9E8}Con &
  \cellcolor[HTML]{FFE9E8}\begin{tabular}[c]{@{}l@{}}Those mass migrants are nothing but low IQ breeders, rugrats and criminals.\end{tabular} \\
  &
  \cellcolor[HTML]{FFE9E8}1.031 &
  \cellcolor[HTML]{FFE9E8}Con &
  \cellcolor[HTML]{FFE9E8}Herding illegals is like herding chickens. It doesn't work without a barrier \\ 
  \multirow{-3}{*}{animal} &
  \cellcolor[HTML]{DBE4FF}0.916 &
  \cellcolor[HTML]{DBE4FF}Lib &
  \cellcolor[HTML]{DBE4FF}\begin{tabular}[c]{@{}l@{}}They’ve told us why they treat immigrants and their children like this.\\They don’t consider them human. They hunt them like animals, they cage them like animals.\end{tabular} \\
  \midrule
 &
  \cellcolor[HTML]{DBE4FF}0.888 &
  \cellcolor[HTML]{DBE4FF}Lib &
  \cellcolor[HTML]{DBE4FF}Wow. Immigrant wifey is on him like a tick. A blood sucker. \#NeverTrump \\
   &
  \cellcolor[HTML]{FFE9E8}0.696 &
  \cellcolor[HTML]{FFE9E8}Con &
  \cellcolor[HTML]{FFE9E8}Deport third world illegals...Leeches on Taxpayers! \\ 
 \multirow{-3}{*}{parasite} &
  \cellcolor[HTML]{DBE4FF}0.683 &
  \cellcolor[HTML]{DBE4FF}Lib &
  \cellcolor[HTML]{DBE4FF}\begin{tabular}[c]{@{}l@{}}They're afraid that those immigrants who tend to congregate in creditor states may bleed off\\the sources of the governmentally funded dole they are now on? Mooches gravy train threatened?\end{tabular} \\ \midrule
 &
  \cellcolor[HTML]{DBE4FF}0.949 &
  \cellcolor[HTML]{DBE4FF}Lib &
  \cellcolor[HTML]{DBE4FF}Ken Cuccinelli Once Compared Immigration Policy To Pest Control, Exterminating Rats \\
 &
  \cellcolor[HTML]{DBE4FF}0.920 &
  \cellcolor[HTML]{DBE4FF}Lib &
  \cellcolor[HTML]{DBE4FF}The president thinks immigrants are an infestation. No subtext here. He literally said they infest the US. \\
\multirow{-3}{*}{vermin} &
  \cellcolor[HTML]{FFE9E8}0.775 &
  \cellcolor[HTML]{FFE9E8}Con &
  \cellcolor[HTML]{FFE9E8}\begin{tabular}[c]{@{}l@{}} I’m sick of paying illegals way for the last 35 years!  The filthy bastards have\\ruined my hometown...thankfully I left SoCal in the 80s before the cockroach infestation.\end{tabular} \\ \midrule
 &
  \cellcolor[HTML]{FFE9E8}1.126 &
  \cellcolor[HTML]{FFE9E8}Con &
  \cellcolor[HTML]{FFE9E8}America Begins to Sink Under Deluge of Illegal Aliens \\
 &
  \cellcolor[HTML]{FFE9E8}1.044 &
  \cellcolor[HTML]{FFE9E8}Con &
  \cellcolor[HTML]{FFE9E8}\begin{tabular}[c]{@{}l@{}}Because right now, those are the very nations pulling up the drawbridge to illegals as their own countries \\ get inundated by vast human waves of Venezuelans flooding over their borders without papers.\end{tabular} \\
\multirow{-3}{*}{water} &
  \cellcolor[HTML]{DBE4FF}0.795 &
  \cellcolor[HTML]{DBE4FF}Lib &
  \cellcolor[HTML]{DBE4FF}\begin{tabular}[c]{@{}l@{}}Or, maybe a high tide raises all boats? There's always a wave of immigrants to the US, \\ and they have all enriched us and made us better.\end{tabular} \\ \midrule
 
 &
  \cellcolor[HTML]{FFE9E8}1.016 &
  \cellcolor[HTML]{FFE9E8}Con &
  \cellcolor[HTML]{FFE9E8}Migrant Caravan Collapses After Pressure From Trump\\
 &
  \cellcolor[HTML]{FFE9E8}0.983 &
  \cellcolor[HTML]{FFE9E8}Con &
  \cellcolor[HTML]{FFE9E8}All that weight from illegal aliens might cause the state to....tip over. \\ 
  
  \multirow{-3}{*}{pressure} &
  \cellcolor[HTML]{DBE4FF}0.940 &
  \cellcolor[HTML]{DBE4FF}Lib &
  \cellcolor[HTML]{DBE4FF} Separating migrant kids from parents could overwhelm an already strained system.\\ \midrule
   &
  \cellcolor[HTML]{FFE9E8}1.427 &
  \cellcolor[HTML]{FFE9E8}Con &
  \cellcolor[HTML]{FFE9E8}They’re importing illegals as replacements. \\
 &
  \cellcolor[HTML]{FFE9E8}1.118 &
  \cellcolor[HTML]{FFE9E8}Con &
  \cellcolor[HTML]{FFE9E8}\begin{tabular}[c]{@{}l@{}}Chicago Mayor-Elect Lori Lightfoot Says She Will Welcome Shipments of Illegals. \\ perfect SEND EM NOW LOAD EM UP AND SHIP THEM TO HER HOUSE ASAP.\end{tabular} \\

\multirow{-3}{*}{commodity} &
  \cellcolor[HTML]{DBE4FF}1.065 &
  \cellcolor[HTML]{DBE4FF}Lib &
  \cellcolor[HTML]{DBE4FF}\begin{tabular}[c]{@{}l@{}}Migrants are a source of \$\$\$\$\$\$\$ for Republican Pals Housing Migrants Is a For-Profit Business. \\ If you take money from people profiting off human misery, you're complicit.\end{tabular} \\ \midrule
 &
  \cellcolor[HTML]{FFE9E8}1.661 &
  \cellcolor[HTML]{FFE9E8}Con &
  \cellcolor[HTML]{FFE9E8}\begin{tabular}[c]{@{}l@{}}Consider the illegals attempting to storm our border as an army of invaders with males using women and\\children as shields. Warn the males they will be targeted by snipers if they attempt to breach our border.',
\end{tabular} \\ 
&
  \cellcolor[HTML]{DBE4FF}1.339 &
  \cellcolor[HTML]{DBE4FF}Lib &
  \cellcolor[HTML]{DBE4FF}Battleground Texas: Progressive Cities Fight Back Against Anti-Immigrant, Right-Wing Forces \\
\multirow{-3}{*}{war} &
  \cellcolor[HTML]{FFE9E8}1.230 &
  \cellcolor[HTML]{FFE9E8}Con &
  \cellcolor[HTML]{FFE9E8}Invaders pillage...send the military. This is a Trojan horse. Democrats want a bloody war at border. \\ \bottomrule
\end{tabular}%
}
\caption{Example liberal and conservative tweets with high metaphor scores for each conceptual domain.}
\label{tab:tweets}
\end{table*}

\subsection{Analysis}
\label{analysis}

\subsubsection{Descriptive Analysis of Scores}
\label{analysis-descriptive}
This section includes descriptive analyses of metaphor scores (§\ref{analysis-descriptive}), results for all regression-based analyses (§\ref{analysis-ideology}, \ref{analysis-engagement}), and example tweets with high metaphor scores (Table \ref{tab:tweets}).

\begin{figure}[htbp!]
    \centering
    \includegraphics[width=\columnwidth]{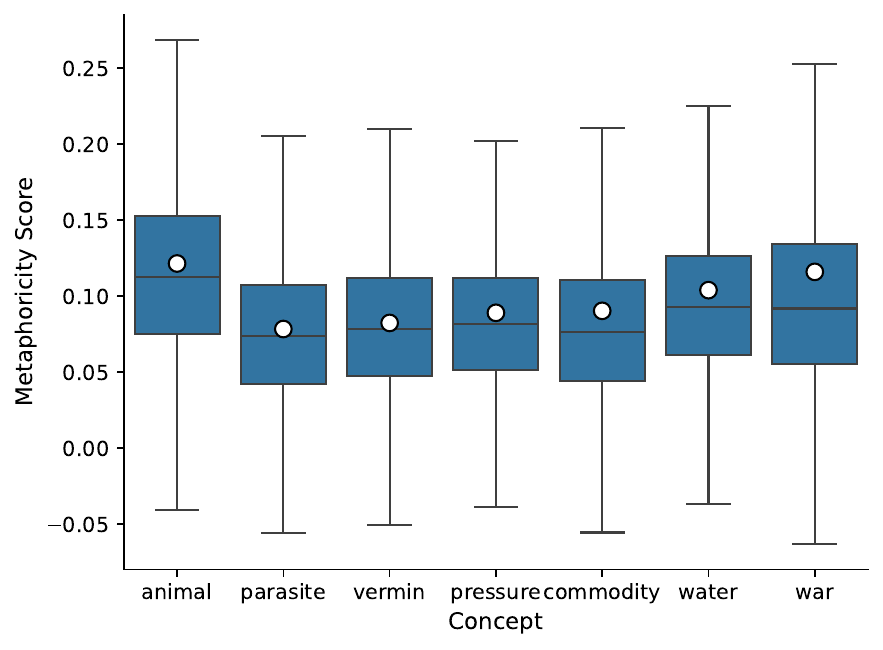}
    \caption{Boxplot showing distribution of metaphor scores for each source domain across all 400K tweets. White dots represent mean scores.}
    \label{fig:scores_boxplot}
\end{figure}

\begin{figure}[htbp!]
    \centering
    \includegraphics[width=\columnwidth]{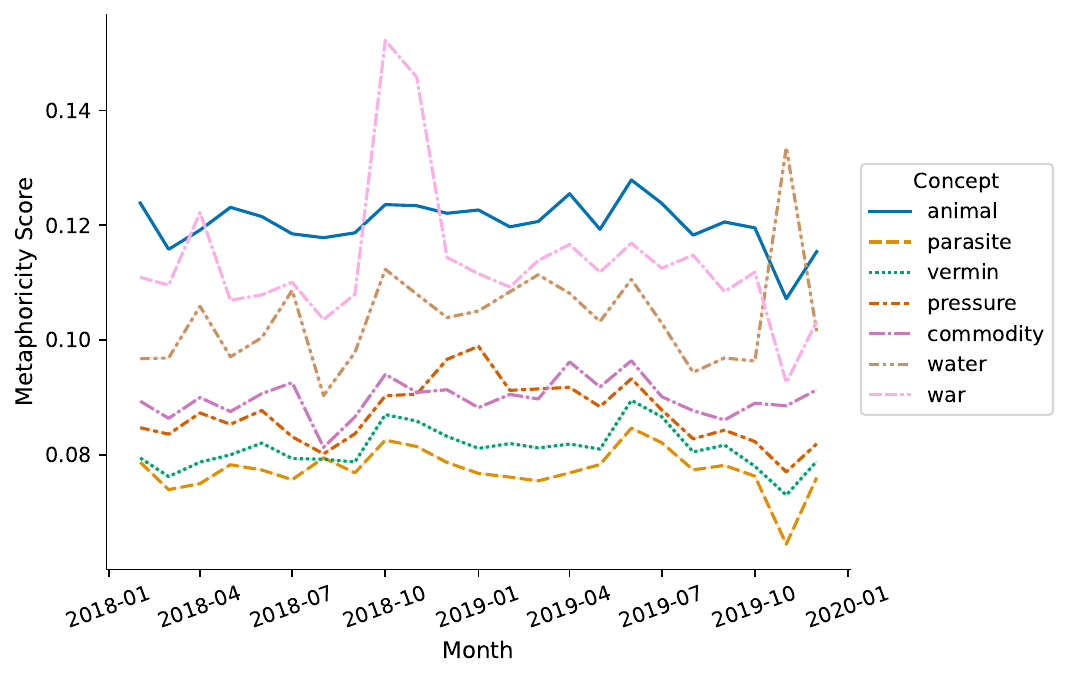}
    \caption{Average metaphor scores by month.}
    \label{fig:scores_by_month}
\end{figure}

\begin{figure}[htbp!]
    \centering
    \includegraphics[width=\columnwidth]{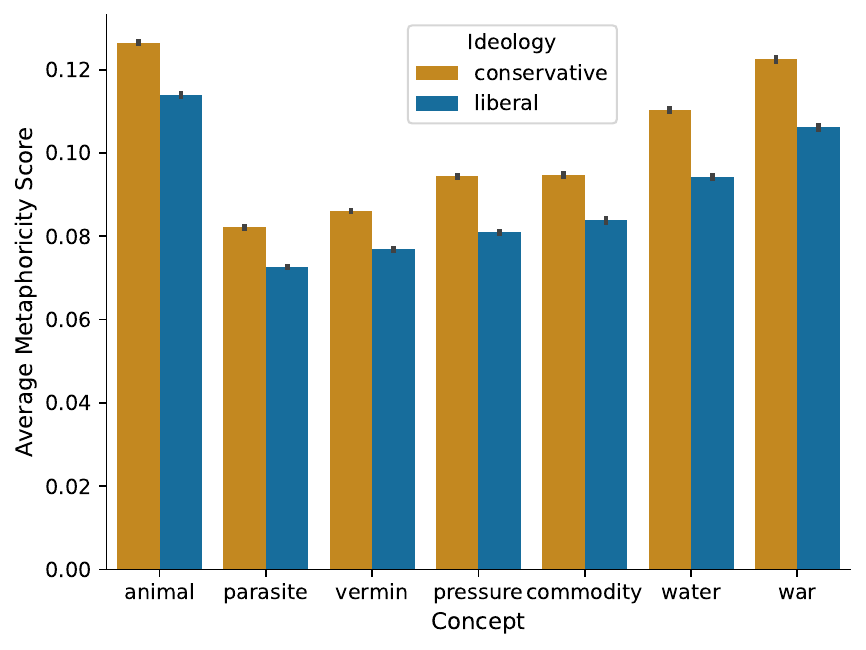}
    \caption{Average metaphor scores of tweets written by liberal and conservative authors for each concept.}
    \label{fig:scores_by_ideology}
\end{figure}

\begin{figure}[htbp!]
    \centering
    \includegraphics[width=\columnwidth]{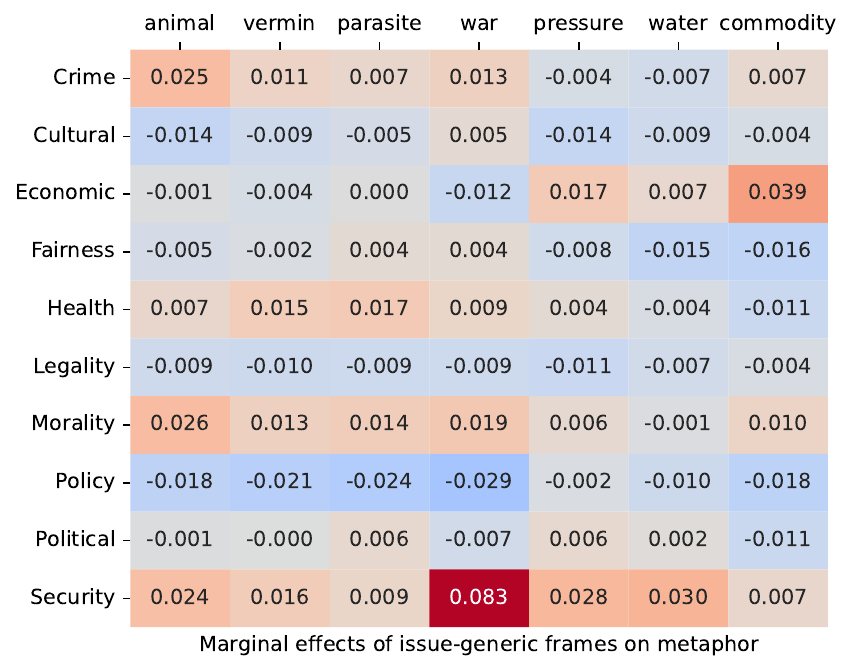}
    \caption{Average marginal effect of issue-generic frames on metaphor scores. Effects are estimated from linear regression models that control for issue-generic frames as fixed effects.}
    \label{fig:frame-heatmap}
\end{figure}

\begin{figure}[htbp!]
    \centering
    \includegraphics[width=.8\columnwidth]{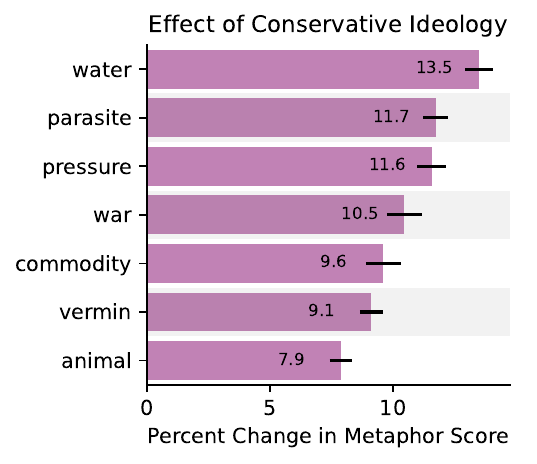}
    \caption{Effect of conservative ideology (relative to liberal) on metaphor score, shown as the average percent change. Percent changes are calculated from marginal effects estimated from regression models that control for issue-generic policy frames as fixed effects.}
    \label{fig:ideology_frames}
\end{figure}

\subsubsection{Role of Ideology in Metaphor}
\label{analysis-ideology}

\begin{figure}[htbp!]
    \centering
    \includegraphics[width=.8\columnwidth]{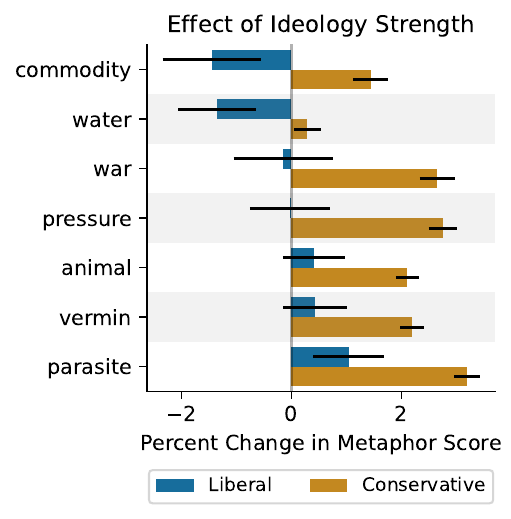}
    \caption{Effect of a one standard deviation increase in ideological strength on metaphor scores. Percent changes are calculated from group-average marginal effects estimated from regression models that control for issue-generic policy frames as fixed effects.}
    \label{fig:strength_frames}
\end{figure}

Figures \ref{fig:ideology_frames} and \ref{fig:strength_frames} shows average marginal effects from regression models that include issue-generic policy frames as fixed effects \citep{mendelsohn2021modeling}. This regression also facilitates analysis of the relationships between issue-generic policy frames, metaphor, and ideology (Figure \ref{fig:frame-heatmap}). Aligning with expectations, some issue-generic frames are strongly associated with particular metaphorical concepts (e.g., \textit{economic} for \textsc{commodity} and \textit{security} for \textsc{war}), and metaphors are more readily used with some frames compared to others (e.g., \textit{security} is more metaphorical than \textsc{cultural identity}). Tables \ref{tab:ideology_effect} and \ref{tab:ideology_effect_frames} show full regression coefficients from models that exclude and include issue-generic frames, respectively.

\begin{table*}[!htbp] \centering 
  \caption{Regression results for the relationship between binary ideology (liberal vs. conservative), ideology strength, and metaphor} 
  \label{tab:ideology_effect} 
\resizebox{\textwidth}{!}{%
\begin{tabular}{@{\extracolsep{-18pt}}lLLLLLLLL} 
\toprule 
\\[-2.8ex] & \multicolumn{1}{c}{animal} & \multicolumn{1}{c}{commodity} & \multicolumn{1}{c}{parasite} & \multicolumn{1}{c}{pressure} & \multicolumn{1}{c}{vermin} & \multicolumn{1}{c}{war} & \multicolumn{1}{c}{water} & \multicolumn{1}{c}{overall}\\ 
\hline \\[-1.8ex]
    ideology & 0.012^{***} & 0.010^{***} & 0.008^{***} & 0.013^{***} & 0.008^{***} & 0.017^{***} & 0.017^{***} & 0.096^{***} \\ 
    strength & 0.001^{***} & -0.002^{***} & 0.002^{***} & -0.001^{*} & 0.001^{***} & -0.0001 & -0.002^{***} & -0.010^{***} \\ 
    ideology:strength & 0.003^{***} & 0.003^{***} & 0.001^{***} & 0.004^{***} & 0.002^{***} & 0.005^{***} & 0.003^{***} & 0.028^{***} \\ 
    \hline
    hashtag & 0.002^{***} & -0.006^{***} & 0.006^{***} & 0.008^{***} & 0.012^{***} & 0.005^{***} & 0.010^{***} & -0.038^{***} \\ 
    mention & -0.004^{***} & -0.002^{***} & -0.004^{***} & -0.004^{***} & -0.002^{***} & -0.011^{***} & 0.006^{***} & -0.024^{***} \\ 
    url & -0.011^{***} & -0.011^{***} & -0.012^{***} & -0.008^{***} & -0.012^{***} & -0.001 & -0.013^{***} & 0.036^{***} \\ 
    quote status & 0.001 & -0.001^{*} & 0.006^{***} & -0.0003 & 0.003^{***} & -0.006^{***} & -0.007^{***} & -0.034^{***} \\ 
    reply & -0.001^{*} & -0.001^{*} & 0.006^{***} & -0.002^{***} & 0.001 & -0.005^{***} & -0.018^{***} & -0.025^{***} \\ 
    verified & -0.009^{***} & -0.004^{***} & -0.010^{***} & -0.007^{***} & -0.009^{***} & -0.013^{***} & 0.002^{*} & -0.036^{***} \\ 
    log chars & -0.017^{***} & -0.022^{***} & -0.016^{***} & -0.006^{***} & -0.015^{***} & -0.020^{***} & -0.001^{**} & -0.002 \\ 
    log followers & -0.0005^{***} & -0.0005^{**} & 0.0001 & -0.0002 & -0.0002 & -0.0003 & -0.004^{***} & -0.007^{***} \\ 
    log following & 0.0003^{*} & 0.0002 & 0.0002^{*} & -0.0002 & 0.0004^{***} & 0.0003 & 0.003^{***} & 0.005^{***} \\ 
    log statuses & 0.001^{***} & 0.001^{***} & 0.001^{***} & 0.001^{***} & 0.001^{***} & 0.002^{***} & 0.004^{***} & 0.011^{***} \\ 
    year:month & -0.000 & 0.000 & 0.000^{***} & -0.000^{***} & 0.000^{***} & 0.000^{***} & -0.000^{**} & 0.000^{***} \\ 
    Constant & 0.205^{***} & 0.200^{***} & 0.156^{***} & 0.116^{***} & 0.153^{***} & 0.192^{***} & 0.083^{***} & 0.005 \\ 
 \hline \\[-1.8ex] 
Observations & \multicolumn{1}{c}{400K} & \multicolumn{1}{c}{400K} & \multicolumn{1}{c}{400K} & \multicolumn{1}{c}{400K} & \multicolumn{1}{c}{400K} & \multicolumn{1}{c}{400K} & \multicolumn{1}{c}{400K} & \multicolumn{1}{c}{400K}  \\ 
R$^{2}$ & 0.033 & 0.027 & 0.052 & 0.020 & 0.052 & 0.024 & 0.028 & 0.041 \\ 
Adjusted R$^{2}$ & 0.032 & 0.027 & 0.052 & 0.020 & 0.052 & 0.024 & 0.028 & 0.041 \\ 
Residual SE & 0.079 & 0.091 & 0.057 & 0.072 & 0.056 & 0.119 & 0.080 & 0.286 \\ 
F Statistic & \multicolumn{1}{r}{961$^{***}$} & \multicolumn{1}{r}{806$^{***}$} & \multicolumn{1}{r}{1,567$^{***}$} & \multicolumn{1}{r}{572$^{***}$} & \multicolumn{1}{r}{1,574$^{***}$} & \multicolumn{1}{r}{689$^{***}$} & \multicolumn{1}{r}{829$^{***}$} & \multicolumn{1}{r}{1,232$^{***}$} \\ 
\bottomrule
  & \multicolumn{8}{r}{$^{*}$p$<$0.05; $^{**}$p$<$0.01; $^{***}$p$<$0.001} \\ 
\end{tabular} }
\end{table*}

\begin{table*}[!htbp] \centering 
  \caption{Regression results for the relationship between binary ideology (liberal vs. conservative), ideology strength, and metaphoricity scores, controlling for issue-generic policy frames.} 
  \label{tab:ideology_effect_frames} 
\resizebox{\textwidth}{!}{%
\begin{tabular}{@{\extracolsep{-18pt}}lLLLLLLLL} 
\toprule 
\\[-2.8ex] & \multicolumn{1}{c}{animal} & \multicolumn{1}{c}{commodity} & \multicolumn{1}{c}{parasite} & \multicolumn{1}{c}{pressure} & \multicolumn{1}{c}{vermin} & \multicolumn{1}{c}{war} & \multicolumn{1}{c}{water} & \multicolumn{1}{c}{overall}\\ 
\hline \\[-1.8ex]
    ideology & 0.009^{***} & 0.008^{***} & 0.009^{***} & 0.009^{***} & 0.007^{***} & 0.011^{***} & 0.013^{***} & 0.077^{***} \\ 
    strength & 0.0005 & -0.001^{**} & 0.001^{**} & -0.0000 & 0.0003 & -0.0002 & -0.001^{***} & -0.006^{***} \\ 
    ideology:strength & 0.002^{***} & 0.003^{***} & 0.002^{***} & 0.003^{***} & 0.002^{***} & 0.003^{***} & 0.002^{***} & 0.021^{***} \\ 
    \hline
    crime & 0.025^{***} & 0.007^{***} & 0.007^{***} & -0.004^{***} & 0.011^{***} & 0.013^{***} & -0.007^{***} & 0.017^{***} \\ 
    cultural & -0.014^{***} & -0.004^{***} & -0.005^{***} & -0.014^{***} & -0.009^{***} & 0.005^{***} & -0.009^{***} & 0.018^{***} \\ 
    economic & -0.001^{***} & 0.039^{***} & 0.0001 & 0.017^{***} & -0.004^{***} & -0.012^{***} & 0.007^{***} & 0.050^{***} \\ 
    fairness & -0.005^{***} & -0.016^{***} & 0.004^{***} & -0.008^{***} & -0.002^{***} & 0.004^{***} & -0.015^{***} & -0.032^{***} \\ 
    health & 0.007^{***} & -0.011^{***} & 0.017^{***} & 0.004^{***} & 0.015^{***} & 0.009^{***} & -0.004^{***} & 0.035^{***} \\ 
    legality & -0.009^{***} & -0.004^{***} & -0.009^{***} & -0.011^{***} & -0.010^{***} & -0.009^{***} & -0.007^{***} & -0.024^{***} \\ 
    morality & 0.026^{***} & 0.010^{***} & 0.014^{***} & 0.006^{***} & 0.013^{***} & 0.019^{***} & -0.001^{**} & 0.001 \\ 
    policy & -0.018^{***} & -0.018^{***} & -0.024^{***} & -0.002^{***} & -0.021^{***} & -0.029^{***} & -0.010^{***} & -0.029^{***} \\ 
    political & -0.001^{*} & -0.011^{***} & 0.006^{***} & 0.006^{***} & -0.0002 & -0.007^{***} & 0.002^{***} & -0.017^{***} \\ 
    security & 0.024^{***} & 0.007^{***} & 0.009^{***} & 0.028^{***} & 0.016^{***} & 0.083^{***} & 0.030^{***} & 0.137^{***} \\ 
    hashtag & 0.002^{***} & -0.004^{***} & 0.007^{***} & 0.007^{***} & 0.012^{***} & 0.003^{***} & 0.009^{***} & -0.040^{***} \\ 
    mention & -0.003^{***} & -0.002^{***} & -0.003^{***} & -0.004^{***} & -0.001^{**} & -0.008^{***} & 0.005^{***} & -0.020^{***} \\ 
    url & -0.013^{***} & -0.010^{***} & -0.014^{***} & -0.009^{***} & -0.014^{***} & -0.004^{***} & -0.014^{***} & 0.032^{***} \\ 
    quote status & 0.0004 & -0.002^{***} & 0.006^{***} & -0.001 & 0.003^{***} & -0.005^{***} & -0.006^{***} & -0.032^{***} \\ 
    reply & -0.001^{**} & -0.002^{***} & 0.005^{***} & -0.001^{**} & 0.0002 & -0.006^{***} & -0.017^{***} & -0.026^{***} \\ 
    verified & -0.005^{***} & -0.003^{**} & -0.007^{***} & -0.005^{***} & -0.006^{***} & -0.009^{***} & 0.003^{***} & -0.029^{***} \\ 
    log chars & -0.019^{***} & -0.022^{***} & -0.017^{***} & -0.011^{***} & -0.014^{***} & -0.024^{***} & -0.001^{*} & -0.017^{***} \\ 
    log followers & -0.001^{***} & -0.0001 & -0.0001 & 0.0000 & -0.0004^{***} & -0.001^{***} & -0.004^{***} & -0.007^{***} \\ 
    log following & 0.0003^{*} & -0.0001 & 0.0001 & -0.0002 & 0.0003^{***} & 0.001^{**} & 0.003^{***} & 0.005^{***} \\ 
    log statuses & 0.002^{***} & 0.001^{***} & 0.001^{***} & 0.001^{***} & 0.001^{***} & 0.002^{***} & 0.004^{***} & 0.011^{***} \\ 
    year:month & 0.000 & 0.000 & 0.000^{***} & -0.000^{***} & 0.000^{***} & 0.000^{***} & -0.000 & 0.000^{***} \\ 
    Constant & 0.203^{***} & 0.203^{***} & 0.156^{***} & 0.135^{***} & 0.146^{***} & 0.199^{***} & 0.085^{***} & 0.057^{***} \\ 
 \hline \\[-1.8ex] 
Observations & \multicolumn{1}{c}{400K} & \multicolumn{1}{c}{400K} & \multicolumn{1}{c}{400K} & \multicolumn{1}{c}{400K} & \multicolumn{1}{c}{400K} & \multicolumn{1}{c}{400K} & \multicolumn{1}{c}{400K} & \multicolumn{1}{c}{400K}  \\ 
$R^{2}$ & 0.102 & 0.076 & 0.125 & 0.069 & 0.137 & 0.124 & 0.063 & 0.092 \\ 
Adjusted $R^{2}$ & 0.102 & 0.076 & 0.125 & 0.069 & 0.137 & 0.124 & 0.063 & 0.092 \\ 
Residual SE & 0.077 & 0.089 & 0.055 & 0.070 & 0.053 & 0.113 & 0.079 & 0.278 \\ 
F Statistic  & \multicolumn{1}{r}{1,887$^{***}$} & \multicolumn{1}{r}{1,369$^{***}$} & \multicolumn{1}{r}{2,375$^{***}$} & \multicolumn{1}{r}{1,236$^{***}$} & \multicolumn{1}{r}{2,637$^{***}$} & \multicolumn{1}{r}{2,365$^{***}$} & \multicolumn{1}{r}{1,129$^{***}$} & \multicolumn{1}{r}{1,686$^{***}$} \\ 
\bottomrule
& \multicolumn{8}{r}{$^{*}$p$<$0.05; $^{**}$p$<$0.01; $^{***}$p$<$0.001}
\end{tabular} }
\end{table*}

\subsubsection{Role of Metaphor in Engagement}
\label{analysis-engagement}

\begin{figure}[t!]
    \centering
    \includegraphics[width=.8\columnwidth]{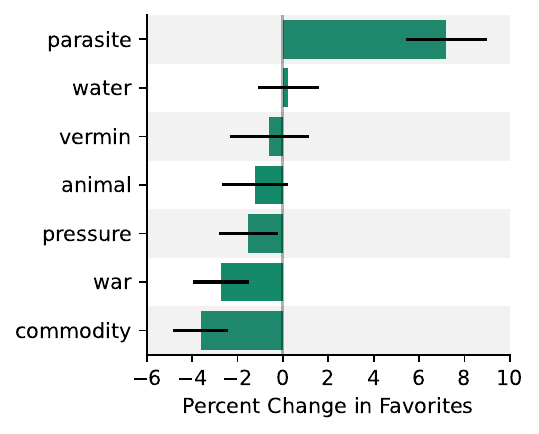}
    \caption{Average marginal effect of metaphor scores on favorites for each concept, estimated from regression models. Effects are shown as percent changes in predicted favorite counts between non-metaphorical and highly-metaphorical tweets (±2 standard deviations).}
    \label{fig:favorite}
\end{figure}

\begin{figure}[t!]
    \centering
    \includegraphics[width=.8\columnwidth]{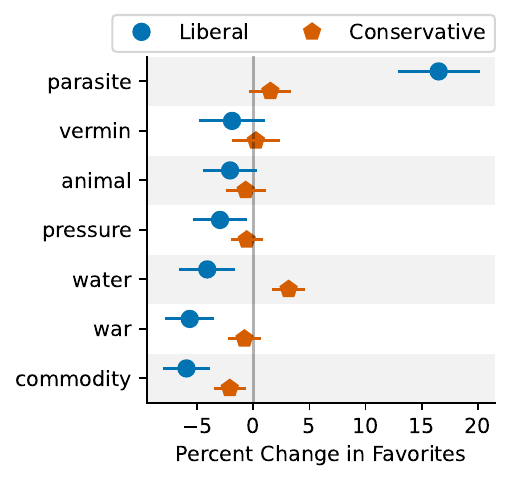}
    \caption{Group-average marginal effects of metaphor on favorites separated by ideology, estimated from regression models. Effects are shown as percent changes in predicted favorite counts between non-metaphorical and highly-metaphorical tweets (±2 standard deviations).}
    \label{fig:favorite_ideology}
\end{figure}

\begin{figure}[htbp!]
    \centering
    \includegraphics[width=.8\columnwidth]{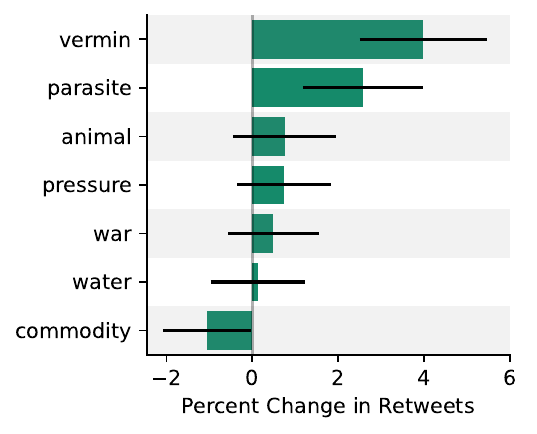}
    \caption{Average marginal effect of metaphor scores on retweets for each concept, estimated from regression models that control for issue-generic frames. Effects are shown as percent changes in predicted retweet counts between non-metaphorical and highly-metaphorical tweets (±2 standard deviations).}
    \label{fig:retweet_frames}
\end{figure}

\begin{figure}[htbp!]
    \centering
    \includegraphics[width=.8\columnwidth]{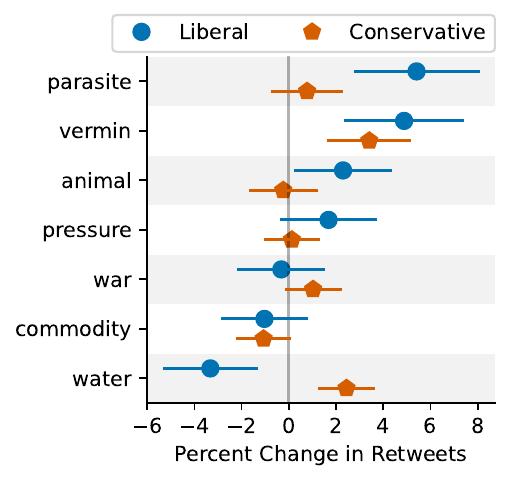}
    \caption{Group-average marginal effects of metaphor on retweets separated by ideology, estimated from regression models that control for issue-generic frames. Effects are shown as percent changes in predicted retweet counts between non-metaphorical and highly-metaphorical tweets (±2 standard deviations).}
    \label{fig:retweet_ideology_frames}
\end{figure}

Figures \ref{fig:favorite} and \ref{fig:favorite_ideology} shows the percent change in favorites associated with a 4 standard deviation change in metaphor score, which corresponds to the difference between a non-metaphorical and a highly-metaphorical tweet. Figure \ref{fig:retweet_frames} illustrates average marginal effects of metaphor on retweets, controlling for issue-generic frames as fixed effects. Figure \ref{fig:retweet_ideology_frames} separates effects on retweets by ideology. All regression coefficients for both sets of models can be found in Tables \ref{tab:engagement}-\ref{tab:engagement_frames}.

\begin{table}[!htbp] \centering 
  \caption{Regression results for the relationship between metaphor scores, ideology, and user engagement (number of favorites and retweets, log-scaled)} 
  \label{tab:engagement} 
\resizebox{\columnwidth}{!}{%
\begin{tabular}{@{\extracolsep{-18pt}}lLL} 
\toprule
\\[-2.8ex] & \multicolumn{1}{c}{favorites} & \multicolumn{1}{c}{retweets} \\  
    \hline \\[-1.8ex]  
    ideology & -0.166^{***} & -0.029^{***} \\ 
    strength & 0.078^{***} & 0.022^{***} \\ 
    ideology:strength & -0.093^{***} & -0.019^{***} \\ 
    \hline
    animal & -0.005 & 0.010^{***} \\ 
    commodity & -0.015^{***} & -0.005 \\ 
    parasite & 0.038^{***} & 0.016^{***} \\ 
    pressure & -0.007 & 0.006 \\ 
    vermin & -0.005 & 0.012^{***} \\ 
    war & -0.015^{***} & 0.0003 \\ 
    water & -0.010^{*} & -0.012^{***} \\ 
    \hline
    animal:ideology & 0.004 & -0.009^{**} \\ 
    commodity:ideology & 0.010^{**} & 0.002 \\ 
    parasite:ideology & -0.034^{***} & -0.012^{**} \\ 
    pressure:ideology & 0.006 & -0.005 \\ 
    vermin:ideology & 0.005 & -0.004 \\ 
    war:ideology & 0.013^{***} & 0.001 \\ 
    water:ideology & 0.018^{***} & 0.016^{***} \\ 
    \hline
    hashtag & -0.048^{***} & -0.019^{***} \\ 
    mention & -0.116^{***} & -0.094^{***} \\ 
    url & -0.307^{***} & -0.163^{***} \\ 
    quote status & -0.011^{*} & -0.071^{***} \\ 
    reply & 0.048^{***} & -0.137^{***} \\ 
    verified & 0.702^{***} & 0.609^{***} \\ 
    log chars & 0.348^{***} & 0.259^{***} \\ 
    log followers & 0.316^{***} & 0.242^{***} \\ 
    log following & -0.120^{***} & -0.087^{***} \\ 
    log statuses & -0.146^{***} & -0.099^{***} \\ 
    year:month &  0.000^{**} & - 0.000 \\ 
    Constant & -0.799^{***} & -0.856^{***} \\ 
 \hline \\[-1.8ex] 
Observations & \multicolumn{1}{c}{400K} & \multicolumn{1}{c}{400K} \\ 
R$^{2}$ & 0.280 & 0.299 \\ 
Adjusted R$^{2}$ & 0.280 & 0.299 \\ 
Residual SE  & 0.872 & 0.701 \\ 
F Statistic  & \multicolumn{1}{r}{5,563$^{***}$} & \multicolumn{1}{r}{6,102$^{***}$} \\ 
\bottomrule
  & \multicolumn{2}{r}{$^{*}$p$<$0.05; $^{**}$p$<$0.01; $^{***}$p$<$0.001} \\ 
\end{tabular} }
\end{table}

\begin{table}[!htbp] \centering 
  \caption{Regression results for the relationship between metaphor scores, ideology, and user engagement (number of favorites and retweets, log-scaled), controlling for issue-generic policy frames.} 
  \label{tab:engagement_frames} 
\resizebox{\columnwidth}{!}{%
\begin{tabular}{@{\extracolsep{-18pt}}lLL} 
\toprule
\\[-2.8ex] & \multicolumn{1}{c}{favorites} & \multicolumn{1}{c}{retweets} \\  
    \hline \\[-1.8ex]  
    ideology & -0.153^{***} & -0.028^{***} \\ 
    strength & 0.074^{***} & 0.020^{***} \\ 
    ideology:strength & -0.088^{***} & -0.017^{***} \\ 
    \hline
    animal & -0.006 & 0.006 \\ 
    commodity & -0.015^{***} & -0.003 \\ 
    parasite & 0.032^{***} & 0.013^{***} \\ 
    pressure & -0.005 & 0.004 \\ 
    vermin & -0.003 & 0.012^{**} \\ 
    war & -0.014^{***} & -0.001 \\ 
    water & -0.006 & -0.008^{*} \\ 
    \hline
    animal:ideology & 0.005 & -0.006^{*} \\ 
    commodity:ideology & 0.008^{*} & -0.0001 \\ 
    parasite:ideology & -0.031^{***} & -0.011^{**} \\ 
    pressure:ideology & 0.005 & -0.004 \\ 
    vermin:ideology & 0.006 & -0.004 \\ 
    war:ideology & 0.014^{***} & 0.003 \\ 
    water:ideology & 0.016^{***} & 0.014^{***} \\ 
    \hline
    crime & -0.003 & 0.029^{***} \\ 
    cultural & 0.029^{***} & 0.005 \\ 
    economic & 0.013^{***} & 0.019^{***} \\ 
    fairness & 0.048^{***} & 0.035^{***} \\ 
    health & -0.012^{***} & 0.018^{***} \\ 
    legality & 0.003 & 0.022^{***} \\ 
    morality & 0.062^{***} & 0.041^{***} \\ 
    policy & -0.002 & 0.011^{***} \\ 
    political & 0.002 & 0.027^{***} \\ 
    security & -0.017^{***} & 0.010^{***} \\ 
    hashtag & -0.048^{***} & -0.021^{***} \\ 
    mention & -0.115^{***} & -0.092^{***} \\ 
    url & -0.304^{***} & -0.165^{***} \\ 
    quote status & -0.016^{**} & -0.074^{***} \\ 
    reply & 0.045^{***} & -0.136^{***} \\ 
    verified & 0.703^{***} & 0.610^{***} \\ 
    log chars & 0.338^{***} & 0.235^{***} \\ 
    log followers & 0.316^{***} & 0.243^{***} \\ 
    log following & -0.121^{***} & -0.087^{***} \\ 
    log statuses & -0.146^{***} & -0.100^{***} \\ 
    year:month &  0.000^{*} &  0.000 \\ 
    Constant & -0.764^{***} & -0.770^{***} \\ 
    \hline \\[-1.8ex] 
Observations & \multicolumn{1}{c}{400K} & \multicolumn{1}{c}{400K} \\ 
R$^{2}$ & 0.281 & 0.300 \\ 
Adjusted R$^{2}$ & 0.281 & 0.300 \\ 
Residual SE  & 0.872 & 0.701 \\ 
F Statistic  & \multicolumn{1}{r}{4,114$^{***}$} & \multicolumn{1}{r}{4,517$^{***}$} \\ 
\bottomrule
  & \multicolumn{2}{r}{$^{*}$p$<$0.05; $^{**}$p$<$0.01; $^{***}$p$<$0.001} \\ 
\end{tabular} }
\end{table}

\begin{table*}[htbp!]
\resizebox{\textwidth}{!}{%
\begin{tabular}{@{}cccccccc@{}}
\toprule
Outcome & Concept & \begin{tabular}[c]{@{}c@{}}Marginal Effect\\ (All)\end{tabular} & \begin{tabular}[c]{@{}c@{}}Percent Change\\ (All)\end{tabular} & \begin{tabular}[c]{@{}c@{}}Marginal Effect\\ (Conservative)\end{tabular} & \begin{tabular}[c]{@{}c@{}}Percent Change\\ (Conservative)\end{tabular} & \begin{tabular}[c]{@{}c@{}}Marginal Effect\\ (Liberal)\end{tabular} & \begin{tabular}[c]{@{}c@{}}Percent Change\\ (Liberal)\end{tabular} \\ \midrule
retweets & animal & 0.005 (0.002) & 0.479 (0.151) & 0.001 (0.002) & 0.130 (0.185) & 0.010 (0.003) & 1.016 (0.257) \\
 & commodity & -0.004 (0.001) & -0.447 (0.127) & -0.004 (0.001) & -0.386 (0.147) & -0.005 (0.002) & -0.541 (0.231) \\
 & parasite & 0.009 (0.002) & 0.875 (0.172) & 0.004 (0.002) & 0.418 (0.190) & 0.016 (0.003) & 1.582 (0.326) \\
 & pressure & 0.003 (0.001) & 0.253 (0.136) & 0.001 (0.001) & 0.054 (0.148) & 0.006 (0.003) & 0.560 (0.259) \\
 & vermin & 0.010 (0.002) & 0.981 (0.182) & 0.008 (0.002) & 0.823 (0.220) & 0.012 (0.003) & 1.225 (0.314) \\
 & war & 0.001 (0.001) & 0.099 (0.130) & 0.001 (0.001) & 0.144 (0.148) & 0.000 (0.002) & 0.029 (0.237) \\
 & water & -0.002 (0.001) & -0.244 (0.138) & 0.004 (0.001) & 0.393 (0.148) & -0.012 (0.003) & -1.214 (0.263) \\ \midrule
favorites & animal & -0.003 (0.002) & -0.305 (0.186) & -0.002 (0.002) & -0.167 (0.229) & -0.005 (0.003) & -0.517 (0.315) \\
 & commodity & -0.009 (0.002) & -0.917 (0.158) & -0.005 (0.002) & -0.525 (0.182) & -0.015 (0.003) & -1.517 (0.284) \\
 & parasite & 0.017 (0.002) & 1.751 (0.216) & 0.004 (0.002) & 0.381 (0.236) & 0.038 (0.004) & 3.892 (0.414) \\
 & pressure & -0.004 (0.002) & -0.382 (0.168) & -0.001 (0.002) & -0.144 (0.184) & -0.007 (0.003) & -0.746 (0.318) \\
 & vermin & -0.002 (0.002) & -0.151 (0.224) & 0.001 (0.003) & 0.060 (0.272) & -0.005 (0.004) & -0.473 (0.384) \\
 & war & -0.007 (0.002) & -0.688 (0.160) & -0.002 (0.002) & -0.195 (0.184) & -0.015 (0.003) & -1.441 (0.290) \\
 & water & 0.001 (0.002) & 0.059 (0.172) & 0.008 (0.002) & 0.780 (0.185) & -0.010 (0.003) & -1.037 (0.328) \\ \bottomrule
\end{tabular}%
}
\caption{Marginal effects of one standard deviation increase in metaphor scores on user engagement, estimated from regression models. Effects for both favorites and retweets are shown, for all tweets as well as group-average marginal effects for liberals and conservatives. Marginal effects correspond to changes in the $ln(x+1)$ scale. Corresponding percent changes in user engagement counts are also shown. Standard errors are displayed in parentheses.}
\label{tab:mfx_engagement_no_frames}
\end{table*}

\begin{table*}[htbp!]
\resizebox{\textwidth}{!}{%
\begin{tabular}{@{}cccccccc@{}}
\toprule
Outcome & Concept & \begin{tabular}[c]{@{}c@{}}Marginal Effect\\ (All)\end{tabular} & \begin{tabular}[c]{@{}c@{}}Percent Change\\ (All)\end{tabular} & \begin{tabular}[c]{@{}c@{}}Marginal Effect\\ (Conservative)\end{tabular} & \begin{tabular}[c]{@{}c@{}}Percent Change\\ (Conservative)\end{tabular} & \begin{tabular}[c]{@{}c@{}}Marginal Effect\\ (Liberal)\end{tabular} & \begin{tabular}[c]{@{}c@{}}Percent Change\\ (Liberal)\end{tabular} \\ \midrule
retweets & animal & 0.002 (0.002) & 0.188 (0.153) & -0.001 (0.002) & -0.059 (0.186) & 0.006 (0.003) & 0.569 (0.258) \\
 & commodity & -0.003 (0.001) & -0.265 (0.132) & -0.003 (0.001) & -0.269 (0.150) & -0.003 (0.002) & -0.260 (0.235) \\
 & parasite & 0.006 (0.002) & 0.638 (0.175) & 0.002 (0.002) & 0.193 (0.191) & 0.013 (0.003) & 1.326 (0.327) \\
 & pressure & 0.002 (0.001) & 0.184 (0.138) & 0.000 (0.002) & 0.032 (0.150) & 0.004 (0.003) & 0.418 (0.261) \\
 & vermin & 0.010 (0.002) & 0.982 (0.184) & 0.008 (0.002) & 0.841 (0.222) & 0.012 (0.003) & 1.199 (0.315) \\
 & war & 0.001 (0.001) & 0.123 (0.134) & 0.003 (0.002) & 0.256 (0.153) & -0.001 (0.002) & -0.080 (0.238) \\
 & water & 0.000 (0.001) & 0.032 (0.140) & 0.006 (0.001) & 0.605 (0.150) & -0.008 (0.003) & -0.842 (0.265) \\ \midrule
favorites & animal & -0.003 (0.002) & -0.259 (0.189) & -0.001 (0.002) & -0.059 (0.231) & -0.006 (0.003) & -0.564 (0.318) \\
 & commodity & -0.010 (0.002) & -1.004 (0.163) & -0.007 (0.002) & -0.689 (0.185) & -0.015 (0.003) & -1.484 (0.288) \\
 & parasite & 0.013 (0.002) & 1.339 (0.219) & 0.001 (0.002) & 0.097 (0.238) & 0.032 (0.004) & 3.277 (0.415) \\
 & pressure & -0.002 (0.002) & -0.199 (0.172) & 0.000 (0.002) & -0.004 (0.187) & -0.005 (0.003) & -0.496 (0.322) \\
 & vermin & 0.000 (0.002) & 0.017 (0.227) & 0.002 (0.003) & 0.241 (0.275) & -0.003 (0.004) & -0.327 (0.386) \\
 & war & -0.005 (0.002) & -0.526 (0.166) & 0.000 (0.002) & 0.021 (0.190) & -0.014 (0.003) & -1.361 (0.293) \\
 & water & 0.003 (0.002) & 0.314 (0.175) & 0.009 (0.002) & 0.933 (0.187) & -0.006 (0.003) & -0.630 (0.331) \\ \bottomrule
\end{tabular}%
}
\caption{Marginal effects of one standard deviation increase in metaphor scores on user engagement, estimated from regression models that additionally control for issue-generic frames. Effects for both favorites and retweets are shown, for all tweets as well as group-average marginal effects for liberals and conservatives. Marginal effects correspond to changes in the $ln(x+1)$ scale. Corresponding percent changes in user engagement counts are also shown. Standard errors are displayed in parentheses.}
\label{tab:mfx_engagement_with_frames}
\end{table*}

\end{document}